\documentclass{article}

\usepackage[numbers]{natbib}
\usepackage[ruled,vlined,linesnumbered]{algorithm2e}
\usepackage{amsmath}
\usepackage{lineno}

\usepackage{PRIMEarxiv}

\usepackage[utf8]{inputenc} 
\usepackage[T1]{fontenc}    
\usepackage{hyperref}       
\usepackage{url}            
\usepackage{booktabs}       
\usepackage{amsfonts}       
\usepackage{nicefrac}       
\usepackage{microtype}      
\usepackage{lipsum}
\usepackage{fancyhdr}       
\usepackage{graphicx}       
\graphicspath{{media/}}     

\title{Parallel Automatic History Matching Algorithm Using Reinforcement Learning
}

\author{
  Omar S. Alolayan \\
  Center for Computational Science and Engineering \\
  Civil and Environmental Engineering\\
  Massachusetts Institute of Technology \\
  Cambridge, MA, USA\\
  \texttt{olyanos@mit.edu} \\
   \And
  Abdullah O. Alomar \\
  Electrical Engineering and Computer Science \\
  Massachusetts Institute of Technology \\
  Cambridge, MA, USA\\
  \texttt{aalomar@mit.edu} \\
  \And
  John R. Williams \\
  Civil and Environmental Engineering\\  Massachusetts Institute of Technology \\
  Cambridge, MA, USA\\
  \texttt{jrw@mit.edu} \\
}

\begin{document}
\maketitle

\begin{abstract}
Reformulating the history matching problem from a least-square mathematical optimization problem into a Markov Decision Process introduces a method in which reinforcement learning can be utilized to solve the problem. This method provides a mechanism where an artificial deep neural network agent can interact with the reservoir simulator and find multiple different solutions to the problem. Such formulation allows for solving the problem in parallel by launching multiple concurrent environments enabling the agent to learn simultaneously from all the environments at once, achieving significant speed up.
\end{abstract}


\section{Introduction}
Optimally developing an oil and gas ﬁeld requires predicting future production using a reservoir model, whose key material properties are tuned in a process called history matching. This process of adjusting the key parameters is non-unique and computationally challenging. Typically, the reservoir model is divided into cells that match the geology of the field.  The key properties of these cells, such as porosity and permeability, are assigned initially using core sample data, where available.  For computational efficiency, the geological model is converted to a reservoir model using upscaling~\mbox{\citep{Durlofsky_2005, Lie_2020,Xian_1996}} to reduce the number of the cells in the model.\\
Due to the challenges of ﬁnding the key properties in each cell, history matching is used to adjust the values of these properties so the model reﬂects historical production data~\citep{Ruijian_2001,Okotie_2018,He_2016}. History matching is typically done by matching the computed pressure and saturation data (oil, gas and water rates) from the simulation model and comparing it the actual historical data. The difference between the actual data and data generated by the reservoir model is then computed using an objective function that quantiﬁes the mismatch between the two quantities. {The problem of history matching can be expressed mathematically as a non-linear least-square optimization problem, where the optimization algorithm minimizes the objective function}:
\begin{equation} \label{eq:1}
{F(u)} = \alpha [ (q - \hat{q})^{T} C_{q}  (q - \hat{q}) + \lambda ({u} - {{u}}^{prior})^{T} C_{{u}} ({u} - {{u}}^{prior})]
\end{equation}

{ where $\alpha$ is a constant scaling factor used scale the quantity of the objective function, $q$ is a vector containing the actual historical pressure or saturation quantity and $\hat{q}$ is a vector that represents the calculated quantity from the simulation model. $C_{q}$ is a weight matrix used to assign weights to production data where it can be used if there is an order of magnitude difference in the production data. ${u}$ is a vector containing the uncertain parameters in the reservoir model. The rates generated by the reservoir simulator $\hat{q}$ can be expressed as a function of the reservoir simulator with respect to ${u}$ i.e., $\hat{q} = f{(u)}$. The second half of the equation is a regularization term where $\lambda$ refers to the regularization parameter, $u^{prior}$ refers to a priori estimate of the uncertain parameters $u$ and $C_{u}$ is another weight matrix that can be used to assign weights for each uncertain parameter.}\\
The history matching problem is an ill-posed problem where multiple solutions can be found to match historical data~\citep{Tomomi_2000,Bruyelle_2019,Li_2018}. Due to such ill-posedness, multiple solutions are used to better assess the uncertainty in the forecasts~\citep{Maschio_2014,Schiozer_2005}.
{The regularization term in the objective function in Equation \mbox{\ref{eq:1}} is used to mitigate the effect of ill-posedness of the history matching problem \mbox{\citep{Ilk_1987,Sayyafzadeh_2012}}.}\\
History matching can be done manually where an experienced professional tunes the parameters of interest to find a matching model. However, manual history matching can be time consuming and prone to human bias and error~\citep{Shahkarami_2012,Arief_2013}.
History matching can also be done with the help of computers, which is known as assisted or automatic history matching. 
In automatic history matching, different methods have been used to search for parameters values that minimize the objective function.
The methods used in the literature include gradient-based, stochastic or probabilistic algorithms.\\
Gradient-based algorithms, such as Levenberg-Marquardt, are exploitative by nature where the algorithm follow the gradient direction in order to search for a minima. Gradient-based algorithms can be powerful as they can convergence quadratically to a local minima under certain conditions~\mbox{\citep{Yamashita_2001}}. However, these algorithms are not feasible for large problems with a large number of parameters~\mbox{\citep{Shirangi_2014}}.\\
Stochastic optimization algorithms such as Genetic Algorithm which is inspired by the biological evolution of genes can better explore the parameters space. By allowing the parameters to evolve independently and then combining the results to choose the model with least error, Genetic Algorithm can handle large number of parameters ~\mbox{\citep{Sanghyun2018}}. However, they require a large number of of function evaluations compared to gradient-based algorithms~\mbox{\citep{Li_2020}}. Probabilistic algorithms, such as Ensemble Kalman Filter (EnKF), uses an ensemble of realizations to represent the model uncertainty where these realizations are updated using a variance minimizing scheme \mbox{\citep{Haugen2008}}. EnKF require fewer function evaluations but may not converge, and the parameters are assumed to have a Gaussian distribution~\mbox{\citep{Li_2020,Sanghyun2018}}.\\
\citet{Li_2020} showed that the history matching problem can alternatively be  solved by  formulating it as a Markov Decision Process and then utilizing  reinforcement learning methods. While~\citet{Li_2020} showed a successful proof of concept, their work was limited to a simple model with a small number of parameters. The scalability and applicability of such approach to a more realistic complex model is yet to be established. This research paper introduces a novel algorithm where the use of a parallel stochastic reinforcement learning policy can efficiently find multiple solutions to the problem using a more complex 3D model and up to 27,000 uncertain parameters. 
Such parallelization allows for utilizing more computing resources to find multiple solutions to large models in a timely manner.\\
Solving the history matching problem in parallel using  deterministic gradient descent algorithms can be complicated as such algorithms require knowing the previous objective function before taking a minimization step.
{Due to the fact that this kind of problems requires a great deal of function evaluations, a lot of research work have been done in order to parallelize the problem or some aspects of it.  In the literature, ensemble Kalman filter was used to solve the history matching problem in parallel~\mbox{\citep{Xian_2006,Xian2007,Lin_2017}}. \mbox{\citet{Tanaka_2018}} developed an optimization workflow in which the field development optimization can be solved in parallel. \mbox{\citet{Sarma_2015}} showed that the model based optimization and uncertainty quantification can be massively parallelized across thousands of commercial cloud computing nodes.}\\
However, to the best of the authors' knowledge, at the time of publishing this paper no previous work has shown that the history matching problem can be solved in parallel using reinforcement learning. In addition, no work has been done previously to show that employing a stochastic policy in reinforcement learning can lead to finding multiple and different solutions to the history matching problem.\\
Section \ref{section:Data} of this paper shows the details of the two reservoir models used as test cases and how their historical data is obtained. 
Then, Section \ref{section:Methodology} explains in details the methodology in which the suggested algorithm is developed and how it can be utilized to find multiple history matched models. 
Next, Section \ref{section:Results} shows the results of applying the algorithm on the two data sets and how the algorithm scales when increasing the number of computing resources. After that, Section \ref{section:Discussion} offers a thorough discussion and analysis of the algorithm and results along with future research work related to improving the algorithm.

\section{Data}\label{section:Data}
\subsection{SPE9}
\begin{figure}
    \centering
    \includegraphics[width=14cm]{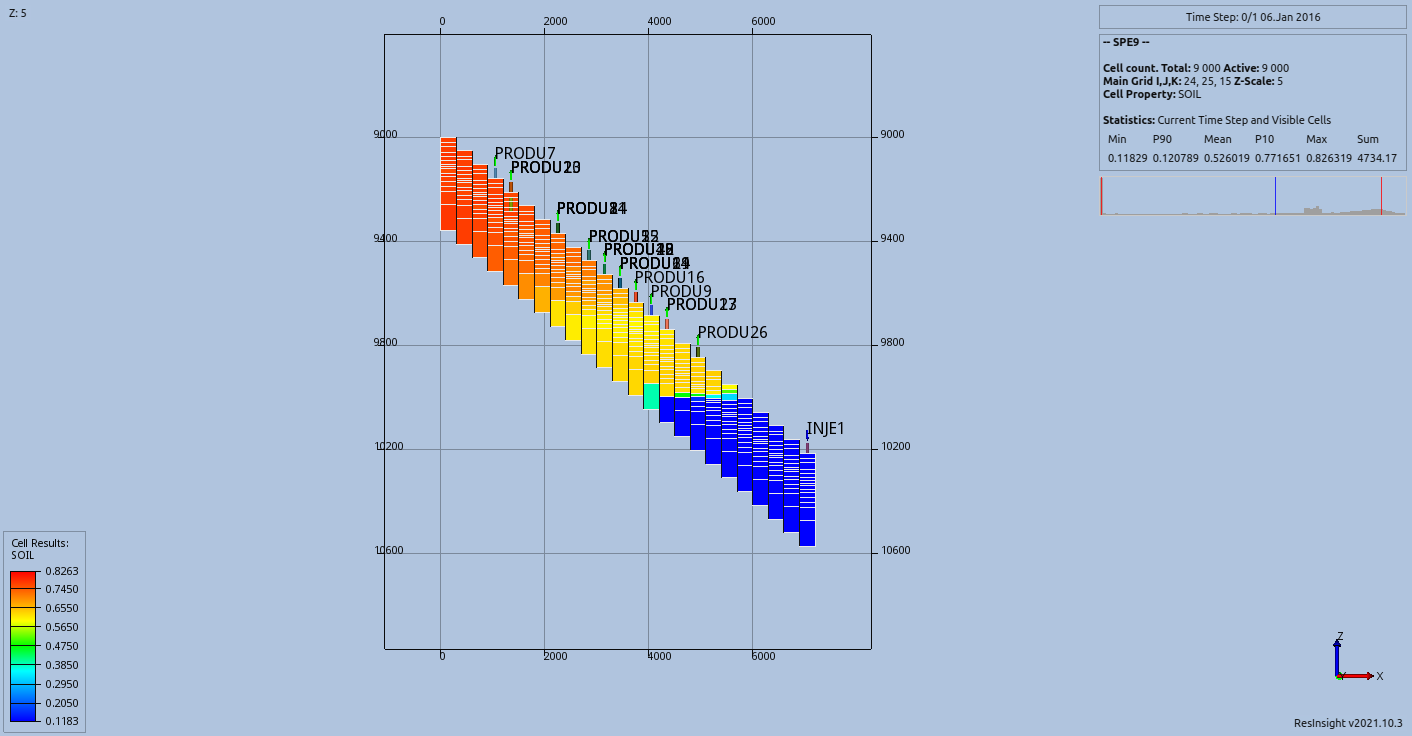}
    \caption{SPE9 reservoir model grid plot. SPE9 reservoir model is a three-dimensional 9000 cells models with 24 cells in the X direction, 25 cells in the Y direction and 15 cells in the Z direction. The reservoir model contains 25 producer wells and one injector well.}
    \label{Fig:SPE9RM1}
\end{figure}

\begin{figure}
    \centering
    \includegraphics[width=14cm]{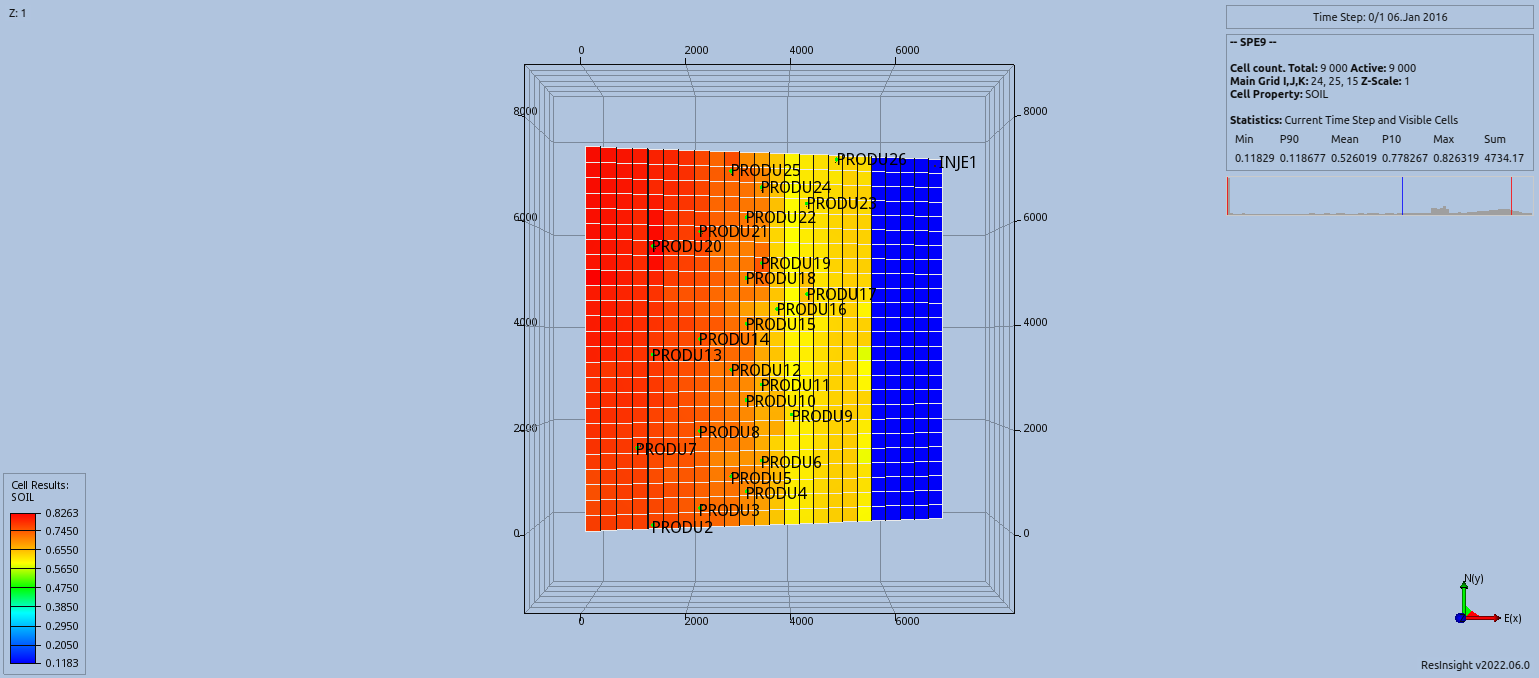}
    \caption{SPE9 reservoir model grid (top view).}
    \label{Fig:SPE9RM2}
\end{figure}

To illustrate the algorithm capacity to find multiple solutions to the history matching problem, SPE9 reservoir model~\citep{SPE_comparative_site,OPM_data_repo} will be used. { The reservoir model developed by~{\citet{Killough_1995}} is a three-dimensional 9000 cells models with 24 cells in the $X$ direction, 25 cells in the $Y$ direction and 15 cells in the $Z$ direction. The reservoir model contains 25 producer wells and one injector well as shown in Figure {\ref{Fig:SPE9RM1}} and {\ref{Fig:SPE9RM2}}.}

The well controls for this model are set as described by~\citet{Killough_1995} and \citet{OPM_data_repo}, where the water injector is set to a maximum rate of 5000 STBW/day with 4000 PSIA as the maximum bottom whole pressure at 9110 ft reference depth. For the producer wells, a 1500 STBO/day is set as the maximum rate at the beginning with a minimum flowing bottom whole pressure of 1000 PSIA at a reference depth of 9110 ft for all the wells.
The model described will be considered as the truth model where the reservoir simulator will simulate this truth model to generate field oil production rate (FOPR) values. { The data generated contains five years worth of FOPR values collected every 15 days.}

\begin{figure}
    \centering
    \includegraphics[width=14cm]{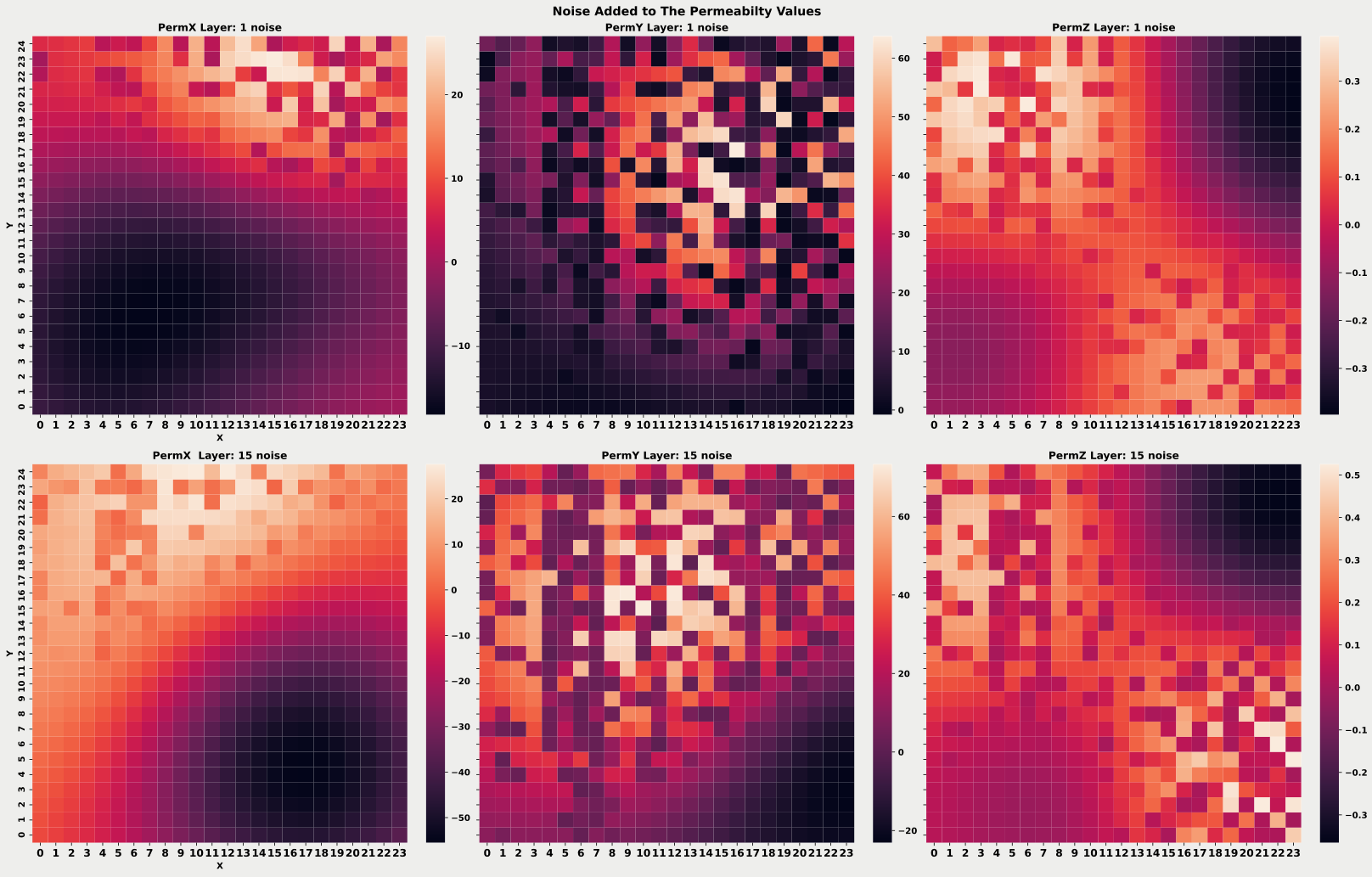}
    \caption{The geo-correlated random noise added to the permeability values in each cell in order to create a synthetic case for the algorithm. The plot shows the first and last layer of the noise added to the truth model permeability values in the X, Y and Z directions.}
    \label{Fig:NoiseSPE9}
\end{figure}

After that, for each cell, the permeability values $K_x$, $K_y$ and $K_z$ are scrambled with a geo-spatially correlated random noise to generate a starting point for the algorithm. The starting point is generated by adding the random noise vector that contains both positive and negative entries to the truth model permeability values. The goal of using a geo-spatial random noise is to ensure that the starting model presents a realistic model where the permeability values are mostly geo-correlated. {Figure {\ref{Fig:NoiseSPE9}} shows the noise added to the permeability values. Since it is difficult to plot all values, the figure only shows a sample of the added noise on the first and last layer of the reservoir model. However, it is important to note that all permeability values across all layers have been scrambled with noise in order to generate the starting point for the algorithm.} 

\begin{figure}
    \centering
    \includegraphics[width=14cm]{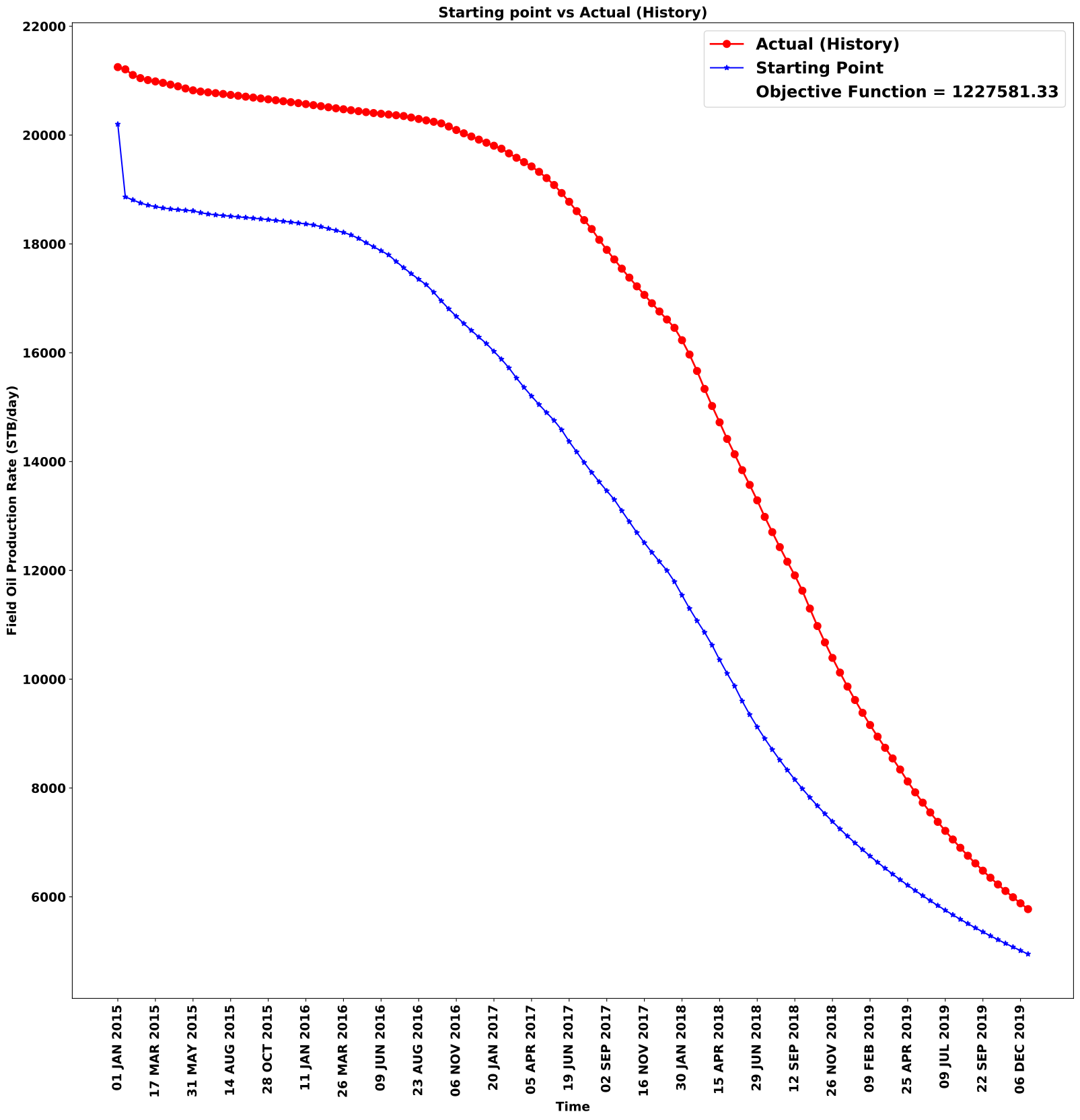}
    \caption{The difference between the actual historical FOPR (Field Oil Production Rate) values and the FOPR values generated from running the simulation on the starting point of the model. The objective function is calculated using Equation \ref{eq:1}.}
    \label{Fig:StartingPointFOPR}
\end{figure}

The truth model is then discarded and only its FOPR values are kept to be considered as the historical values $q$. The starting point after scrambling the truth values is shown in Figure \ref{Fig:StartingPointFOPR}. The goal of starting with a known model then scrambling its permeability values is to imitate a real life scenario where the actual uncertain values (in this case permeability values) are not known exactly and the historical pressure or saturation data (in this case FOPR) are used to tune the uncertain parameters. 

\subsection{SPE1}
To measure the scalability of the suggested algorithm, a smaller reservoir model is used. The main reason for choosing a smaller model is the fact that the scalability test requires repeating the experiment for a number of trials as it will be discussed in the Methodology Section. 

\begin{figure}
    \centering
    \includegraphics[width=14cm]{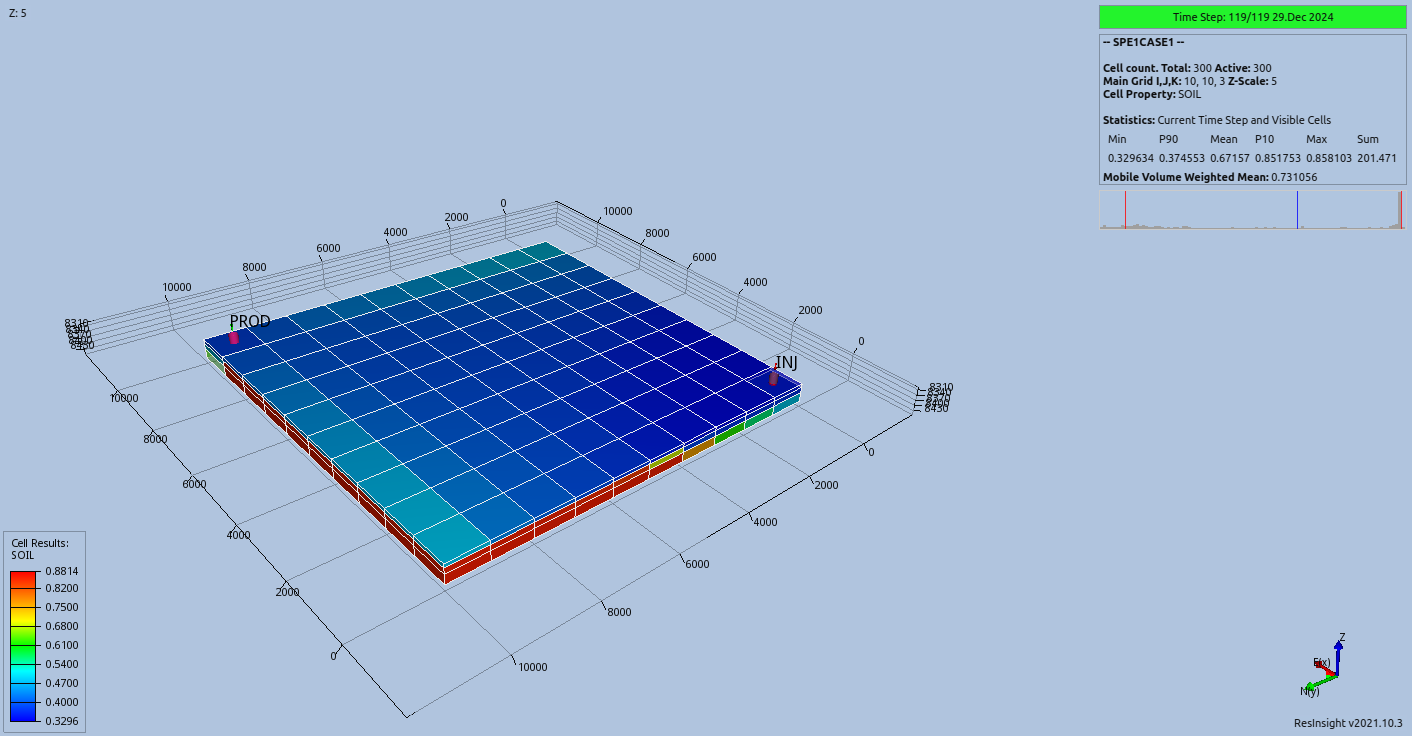}
    \caption{SPE1 reservoir model grid. SPE1 reservoir model is small 3D model with 10 cells in the X direction, 10 cells in the Y direction and 3 cells in the Z direction.The reservoir model has one injector located in cell (x = 1, y = 1,z = 1) and one producer located in cell (x = 10, y = 10,z = 1) where the measurement of BHP is recorded.}
    \label{Fig:SPE1RM}
\end{figure}

\indent The model used for testing the scalability in this research paper is a three-dimensional 300 cells model with 10 cells in the $X$ direction, 10 cells in the $Y$ direction and 3 cells in the $Z$ direction as shown in in Figure \ref{Fig:SPE1RM}. The reservoir model has one injector located in cell ($x=1$, $y=1$,$z=1$) and one producer located in cell ($x=10$, $y=10$,$z=1$) where the measurement of BHP is considered in the objective function.

{The well controls for this model are set as described by~{\citet{Odeh_1981,OPM_data_repo}}, where the gas injector (INJ) is set to a maximum rate of 100 MMscf/day with 9014 PSIA set as the maximum bottom whole pressure. For the producer well (PROD), a 20,000 STBO/day is set as the maximum rate with a minimum flowing bottom whole pressure of 1000 PSIA.}

\begin{figure}
    \centering
    \includegraphics[width=14cm]{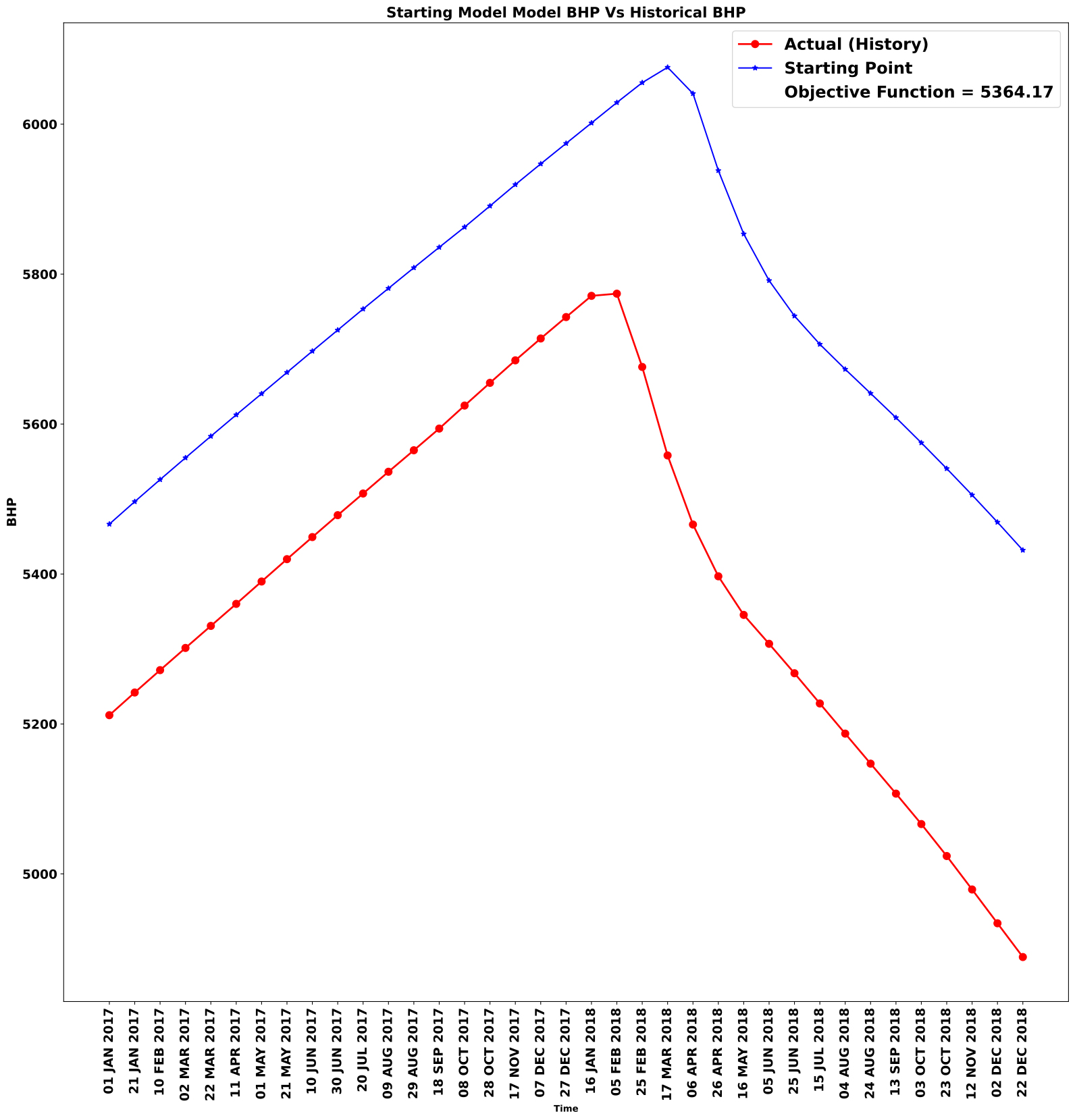}
    \caption{The difference between the actual historical BHP (Bottom Hole Pressure) values and the BHP values generated from running the simulation on the starting point of the model. The objective function is calculated using Equation \ref{eq:1}.}
    \label{Fig:StartingPointBHP}
\end{figure}

The second data set, shown in Figure \ref{Fig:StartingPointBHP}, employs a synthetic historical data obtained from running SPE1~\citep{Odeh_1981,SPE_comparative_site,OPM_data_repo} by following the same procedure used in the first data set, except that in this data set the Bottom Hole Pressure (BHP) values for the producer well is used to match the history instead of FOPR.

\section{Methodology}\label{section:Methodology}
\subsection{Reinforcement Learning}
\begin{figure}
    \centering
    \includegraphics[width=14cm]{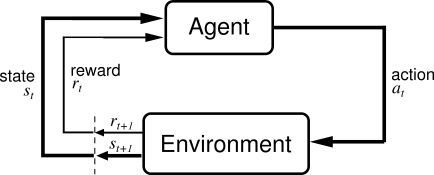}
    \caption{~\citet{Sutton_2018} illustration of how the artificial deep neural network agent learns by interacting with the environment. The agent reads the current state of the environment, takes an action and collects a reward for the action taken.}
    \label{Fig:MDPinRL}
\end{figure}

{In recent years, machine learning has been employed extensively to tackle research problems thanks to its capacity to map input data to a useful output by identifying certain features. In the literature, machine learning tasks usually falls into one of three categories: supervised learning, unsupervised learning and reinforcement learning}~\citep{Lapan_2018}. {In supervised learning, a model is trained with labeled data i.e., examples in order to later predict new data with no label. In unsupervised learning, the model is trained with unlabeled data points to identify certain features  and subsequently classify them into clusters. In reinforcement learning,  the model learns a policy that maximizes a certain reward in an environment of interest. The policy is often learned by directly  interacting with the environment}~\citep{Hammoudeh_2018}.\\
In the oil and gas industry, machine learning have been used extensively in a wide variety of problems.~\citet{Montgomery_2020} used supervised learning to build data driven models to forecast shale gas production where~\citet{Zhang_2019} developed a multi-component method for reservoir characterisation using unsupervised learning.~\citet{Miftakhov_2020} used reinforcement learning to maximize the Net Present Value (NPV) of waterflooding by training a reinforcement learning agent to control the water injection rate.\\
Reinforcement learning is somewhat similar to the way humans learn by interacting with an environment and using a trial and error mechanism.  {The reinforcement learning agent, as shown in Figure} \ref{Fig:MDPinRL}{, observes the state of the environment, takes an action and collects a reward for its action. The agent will learn from these interactions how to search the parameters space in order to maximize its rewards.  Reinforcement learning gained more popularity after} ~\citet{Hasselt_2015} {showed that a reinforcement learning agent was capable of achieving super human intelligence just by monitoring pixels on the screen.}
However,  reinforcement learning is more generic, and is suitable for all problems that can be formulated a Markov Decision Process (MDP). \\
An MDP represents the sequential decision making paradigm in which the actions taken by the agent does not only affect the immediate reward but it also affects future long term rewards and future states~\citep{Sutton_2018}. {MDPs are often defined by the tuple $(\mathcal{S}, \mathcal{A},  \mathcal{P}, \mathcal{R}, \gamma)$, where $\mathcal{S}$ denotes the state space,  $\mathcal{A}$ denotes the action space,  $\mathcal{P}: \mathcal{S} \times \mathcal{A} \times \mathcal{S} \to [0,1] $ denotes the  transition kernel, $\mathcal{R} : \mathcal{S} \times \mathcal{A}  \to \mathbb{R}$ denotes the reward function, and $\gamma$ denotes the discount factor. For the transition kernel, we use $\mathcal{P}(s'\mid s, a)$ to denote the probability of transitioning from state $s$ to state $s'$ if action $a$ is taken. }
{We denote a stochastic policy by $\pi: \mathcal{S} \times \mathcal{A}  \to [0,1]$, where we use $\pi(a\mid s)$ to denote the probability of choosing action $a$ when in state $s$ under policy $\pi$.
Under each policy, the value function $V_\pi(s)$ represents the reward-to-go when starting from state $s$ and using policy $\pi$. Specifically, we have}
\begin{align}
    V_{\pi}\left(s_{t}\right)=\mathbb{E}\left[\sum_{l=0}^{\infty} \gamma^{l} R\left(s_{t+l}, a_{t+l}\right)\right], 
\end{align}
{
where $s_{t+1} \sim \mathcal{P}(s_{t+1}\mid s_t, a_t)$, and  $a_{t} \sim \pi(a_t\mid s_t)$, and the expectations are naturally taken with respect to the randomness in the policy and the transition kernel.  In reinforcement learning, the goal is often to estimate the optimal policy $\pi^*(s) = \arg \max_\pi V_{\pi}\left(s \right)$ or the optimal value function $V^*(s)  =\max_\pi V_{\pi}\left(s \right)$. }

\subsection{History Matching Using Reinforcement Learning}

\begin{figure}
    \centering
    \includegraphics[width=14cm]{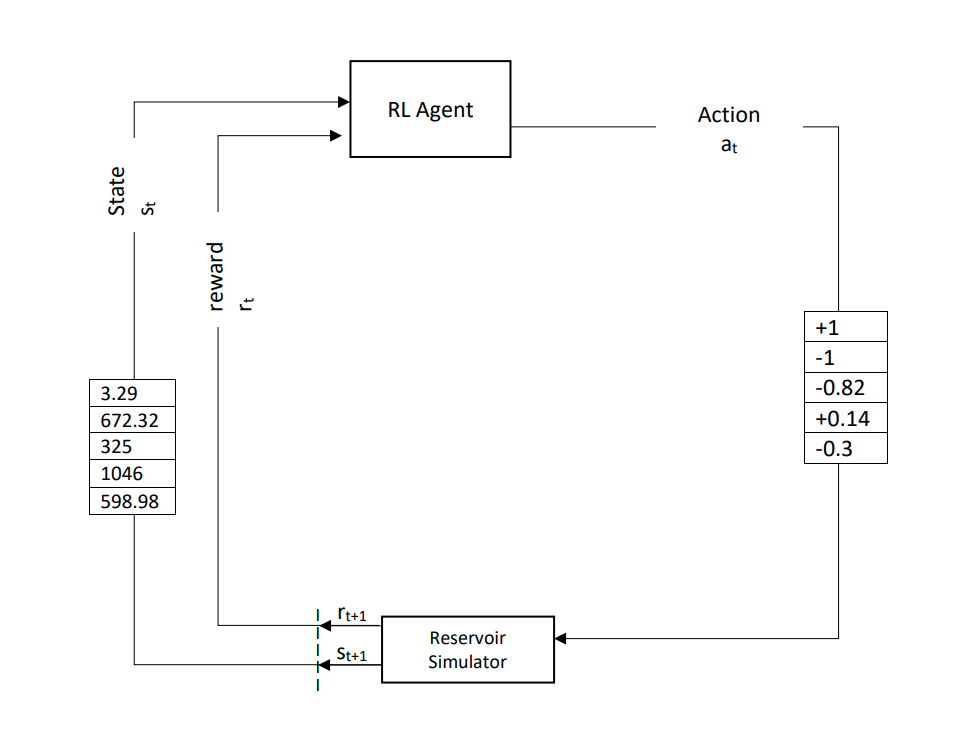}
    \caption{Reformulating the history matching problem from a least-square mathematical optimization problem into a Markov Decision Process by creating an environment that allows the agent to interact with the reservoir simulator. The agents observes the current state $s_t$ of the uncertain parameters ${u}$, takes action $a_t$ against these parameters and collects a reward $r_t$ quantifying the quality of its action. }
    \label{Fig:RSinRL}
\end{figure}

In this paper, we utilize reinforcement learning to solve the history matching problem by formulating it into a Markov Decision Process as illustrated in Figure \ref{Fig:RSinRL}. By using the reservoir simulator as an environment where the {reinforcement learning} agent $Ag$ can take action $a_t$ and receive the new state of the environment $s_t$ along with a reward $r_t$ quantifying how good the action taken by the agent is. The current state returned by the environment is a vector containing all the uncertain parameters that need to be tuned in order to match the history and is equivalent to ${u}$ in Equation \ref{eq:1}.\\
For both cases used in this research paper, $K_x$, $K_y$  and $K_z$ permeability values of each cell are stacked into a vector as follows:
\begin{linenomath*}
\begin{align}
    {u} = \begin{bmatrix}
           K_x (x=0, y=0, z =0)\\
           \vdots \\
           K_x (x=X, y=Y, z =Z) \\
           K_y (x=0, y=0, z =0)\\
           \vdots \\
           K_y (x=X, y=Y, z =Z) \\
           K_z (x=0, y=0, z =0) \\           
           \vdots \\
           K_z (x=X, y=Y, z =Z) \\            
         \end{bmatrix}
  \end{align}
 \end{linenomath*}
 
where this vector is considered as the state of the environment $s_t$. The action $a_t$ taken by the agent is also a vector with the same dimensions as the state, allowing the artificial deep neural network agent to act on the uncertain parameters and adjusting them to find a solution. At each reinforcement learning time-step $t$, the action taken by the agent is given a reward value $r_t$ to quantify the quality of the action allows the agent to learn whether the action it took is good or bad. In this work, Open Porous Media Flow reservoir simulator~\citep{OPM_Paper} is used in the reinforcement learning environment.
\\
Formulating the history matching problem in such manner allows for the use of reinforcement learning, where the agent will be able to learn from the reservoir simulator how to take actions that lead to minimizing the objective function. 
The agent is allowed to only select actions that are within a certain range i.e., the agent action space $\mathcal{A}$ at each entry of the action vector is = [-$K_\Delta$,+$K_\Delta$]. It is important to note that $K_\Delta$ is a design parameter that can be chosen by the engineers running the experiment based on their knowledge of the reservoir model at hand. The choice of $K_\Delta$ must not be too small nor too large. Choosing a small $K_\Delta$ value will change the objective function by a very small quantity, causing very slow convergence. While, choosing a large value can encourage the agent to take actions that push the permeability values out of allowed bounds.\\
If the agent takes an action that causes the current state $s_t$ to be out of bounds (outside of all possible state space $\mathcal{S}$), then the current state is clipped to satisfy physical restrictions. For example, if the agent takes an action that results in negative permeability values in some of the vector entries, then these values are set to zero in order to avoid feeding the reservoir simulator negative permeability values causing it to crash.
The $K_\Delta$ parameter is similar to the step size parameter in gradient descent optimization algorithms where selecting a very small value will slow the convergence process and a big value may cause divergence. {Another approach can be used to avoid reaching states with negative permeability values is to change the actions of the agent to multiply the current permeability value instead of addition or subtraction. By setting $K_\Delta$ to a small value, to avoid large updates, for example 0.1 the agent can sample an action between $1+K_\Delta$  and $1-K_\Delta$ in order to tune the permeability.}\\
Reinforcement learning process consist of episodes where an episode is defined as a series of reinforcement learning time-steps starting from the initial state and ending when a termination criteria is met. This termination criteria is usually governed by a reward function.
Typically, in reinforcement learning the reward function is designed in manner where the agent is given a positive reward for good desired actions and a negative reward for bad undesired actions. However, choosing a reward function can be a challenging task as sometimes there are good actions and even better actions or bad actions and worse actions which can be difficult to quantify. Luckily, in the history matching problem, the quality of the action can be measured directly from the objective function, where the difference between the value of the previous objective function and the value of the new objective function after the agent took an action can quantify the quality of the match.\\
In the context of applying reinforcement learning to solve the history matching problem, the reward function can be computed as:
\begin{linenomath*}
\begin{equation} \label{RewardFunc}
r_t = \mathcal{F}_{t-1} - \mathcal{F}_{t}
\end{equation}
\end{linenomath*}
$\mathcal{F}_{t-1}$ is computed using Equation \ref{eq:1} and refers to the value of the objective function at time = ($t-1$) before the agent takes an action, whereas $\mathcal{F}_{t}$ refers to the value of the objective function after the agent takes an action. This reward function will return a positive value if the agent took a good action that led to reducing the objective function and a negative value if the agent took a bad action that caused the objective function to become larger. The advantage of using this reward function is that it will assign larger rewards to actions that leads to a larger reduction in the objective function, thus helping to guide the agent to take actions that will speed up the convergence process. Assigning a higher reward to an action increases the probability of choosing that action again given the same state.
{The value of the objective function is scaled down using a scaling factor $\alpha$ in order to avoid feeding the deep neural networks very large quantities that might destabilize the training process.\\
The choice of the error quantification function can affect the reward function and subsequently the performance of the algorithm. Equation ~\mbox{\ref{eq:1}} utilizes a weighted squared error formula for the objective function. The weighted squared error is used since it is the common choice for history matching problems and it would provide the agent with a good feedback that quantify the quality of its action. However, weighted root squared error might provide a better error quantification for the algorithm as it would provide a somewhat uniform reward function to the agent unlike squared error, which penalizes larger error more severely possibly leading to a non-uniform reward function.}\\
The reward function defined in Equation \ref{RewardFunc} is not enough on its own and it needs to address termination criteria when the agent takes an action that reduces the objective function to the tolerance level accepted. In order to address that, in this research paper, the agent is assigned a big constant reward (i.e., 50,000) to incentivize it with a big reward upon good episode termination.
In addition, in order to save computing power and incentivize the agent to take actions that speed up the convergence process, the environment sets a time limitation~\citep{Pardo_2018} on the maximum time-steps allowed per episode. Once the maximum time-steps is reached without meeting any termination criteria, the episode terminates with a constant negative reward (i.e., -20,000) if the agent does not find a solution. The maximum time-steps per episode parameter is also a design parameter and should be chosen based on the problem at hand but it must not be too small. Small time-steps per episode can hinder the agent from exploring the environment properly due to the fact that in reinforcement learning the agent sometimes might pick actions that are not good at the current time-step, but can lead to better rewards in the future time-steps. \\
Furthermore, if the agent diverges away from the solution by making a sequence of bad decisions rendering the objective function to grow large. In this case, the episode is terminated with a big negative constant rewards (i.e., -50,000) as soon as the objective function value becomes twice as large as the initial objective function. In addition to saving computing power, such a limitation on the value of the objective function will punish the agent so it can learn from that mistake and avoid following such a trajectory in the future.
The rewards of termination criteria are also design parameters and the engineers running the experiment should choose proper values relevant to the problem at hand. The rewards should be chosen so they can incentives the agent towards meeting the tolerance criteria as well as providing a good feedback to the engineers so they can monitor the learning process.
{This formulation of the history matching problem into a Markov Decision Process  means that at each reinforcement learning time-step, a simulation run is required to compute the new objective function and assess the quality of the action taken by the agent}.

\subsection{Proximal Policy Optimization}
\begin{figure}
    \centering
    \includegraphics[width=10cm]{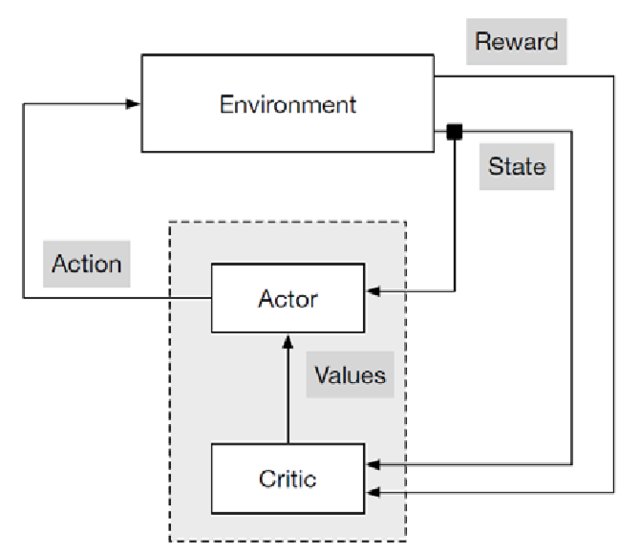}
    \caption{ \citet{Diederichs2019} illustration of actor-critic architecture.}
    \label{Fig:ActorCritic}
\end{figure}

\begin{figure}
    \centering
    \includegraphics[width=14cm]{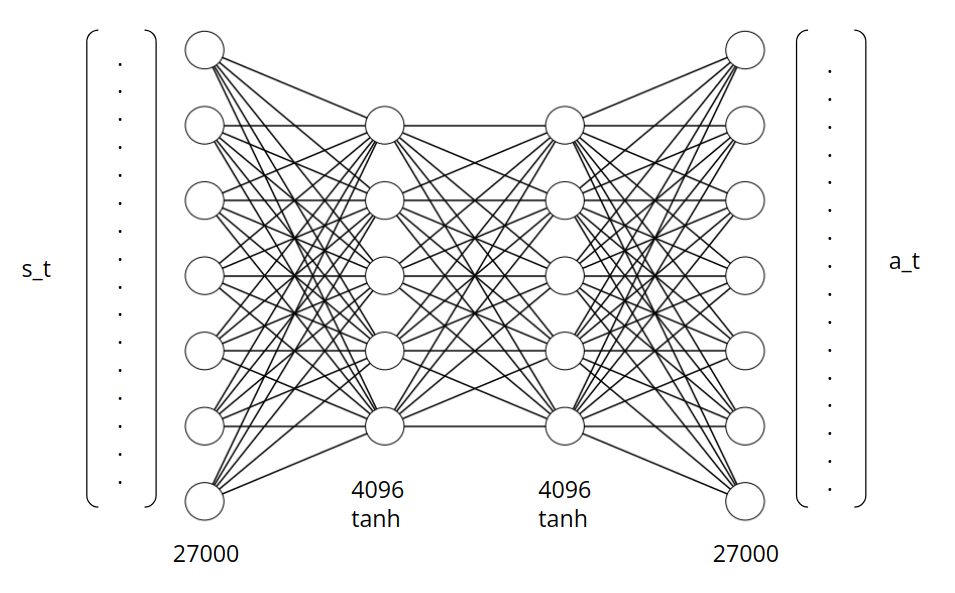}
    \caption{The deep neural network design for the actor network ${\pi_{\theta}}$ is composed of an input layer containing 27,000 neurons allowing it to observe the current state of each uncertain parameter, two hidden layers where each layer contains 4096 tanh activated neurons and an output layer that also equal to 27,000 allowing it to take action against 27,000 different uncertain values.}
    \label{Fig:ActorNet}
\end{figure}

\begin{figure}
    \centering
    \includegraphics[width=14cm]{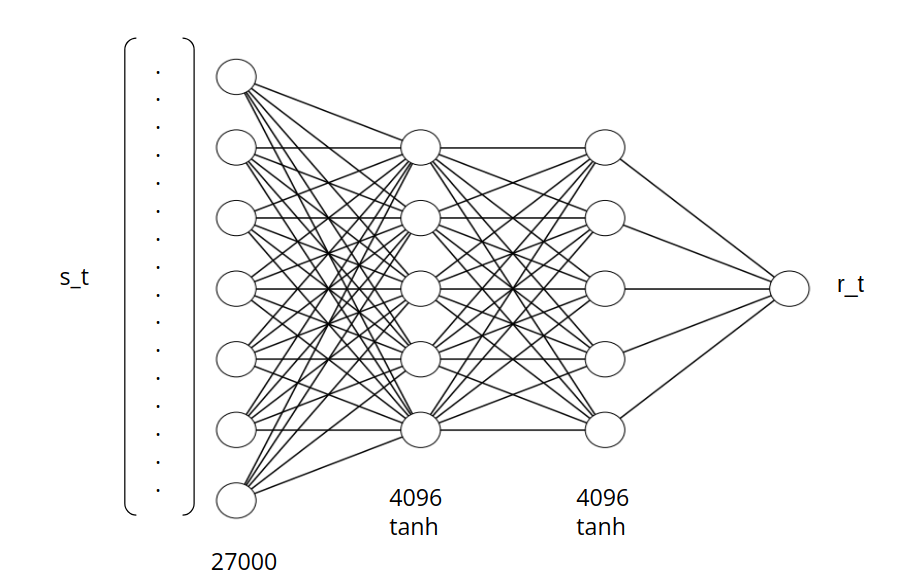}
    \caption{The deep neural network design for the critic network $V(\theta)$ is composed of an input layer containing 27,000 to observe the current state of the environment, two hidden layers where each layer contains 4096 tanh activated neurons and an output layer that also equal to one neuron as it only needs to estimate one value.}
    \label{Fig:CriticNet}
\end{figure}

In this work, we use Proximal Policy Optimization (PPO)~\citep{Schulman_2017} algorithm to find multiple solutions to the history matching problem.{ 
PPO algorithm is an actor-critic method in which the agent utilizes two deep neural networks  parameterized by $\theta$, one for the actor and one for the critic as shown in Figure \mbox{\ref{Fig:ActorCritic}}. The value function and the policy are represented by these two deep neural networks. Specifically, the actor and critic networks represent the policy and the value function, respectively.}
\\
One of the prevalent issues in policy gradient methods is the destructively large policy updates. To overcome this issue, PPO suggests a simple surrogate objective that penalizes large changes in the policy update. 
Specifically, the objective function in PPO is as follows
\begin{linenomath*}
\begin{equation} \label{eq:2}
L_t (\theta) = \mathbb{E} [ L^{clip}_{t} (\theta) - c_{1} L_{t}^{VF}(\theta)],
\end{equation}
\end{linenomath*}

where $\mathbb{E}$ refers to the expectation operator, $L^{clip}_{t} (\theta)$ represents the clipped objective function for the policy network and $L_{t}^{VF}$ represents the error in estimating the value function. {Specifically, for the policy loss, the clipped objective function is defined as }
\begin{linenomath*}
\begin{equation} \label{eq:TBA}
{L_t} ^{clip} {(\theta)} = \mathbb{E}_{t} [ \min( \delta_{t}{(\theta)} {\hat{A}_{t}}, \text{clip}(\delta_{t} (\theta), 1-e , 1+e)) ],
\end{equation}
\end{linenomath*}
{where $\delta_{t} (\theta)$ is a probability ratio of the new policy to the old policy computed as $\frac{\pi_{\theta}(a_t|s_t)}{\pi_{{\theta}_{old}}(a_t|s_t)}$,  $e$ is a clipping range variable and is set to 0.2 in the case study, and $\hat{A}_t$ represents the estimate of the advantage function calculated as 
}
\begin{linenomath*}
\begin{equation} \label{eq:4}
\hat{A}_t = \sum_{k=0}^{T} {\gamma}^k r_{t+k} - V_\theta(s_t), 
\end{equation}
\end{linenomath*}
{where $V_\theta(s_t)$ is the value function estimate given by the value network. Note that the advantage function at time $t$ is defined as the discounted sum of rewards (starting from t) minus a baseline. The value function at $s_t$ is often used as the baseline, as done in PPO. The goal of the clipping operator is to stop drastic policy updates based on the noisy estimation $\hat{A}_t$, and hence it avoids drastic policy updates.
}
\indent {Further, $L_{t}^{VF}(\theta)$, which denotes the loss in estimating the value function is defined as}
\begin{linenomath*}
\begin{equation} \label{eq:5}
L_{t}^{VF} (\theta) = ( V_{\theta}(s_t)  - V_t)^2 = \hat{A}^2_t ,
\end{equation}
\end{linenomath*}
where $V_t = \sum_{k=0}^{T} {\gamma}^k r_{t+k} $ are samples estimates collected from the environment, and $T$ is the maximum number of steps. Recall that the value function $V_\theta(s_t)$ represents an \emph{estimate} of the expected rewards for a given state $s_t$ at a given time = $t$. This estimate is estimated using the critic network, which along with the policy network, is updated during training. Recall also that $\gamma$ is a discount factor which controls the trade off between short-term and long-term rewards. Specifically,  as $\gamma$ decreases, more emphasis is given to  short-term rewards.\\
{{While seemingly complicated, the objective function described above is quite intuitive. Minimizing}  $L_{t}^{VF}$ {ensures that the critic's estimate of the value function is close to what the observed samples suggest. On the other hand, maximising} ${L_t} ^{clip}$, {if we ignore the clipping, encourages policies with a better advantage, i.e., policy that chooses relatively better actions. The clipping, as mentioned above, mitigate potentially  destructive updates due to potentially huge values of  }$\delta_t$}.\\
Note that while PPO is used in this work, other actor-critic methods such as A2C~\citep{Volodymyr_2016} which supports sampling from parallel environments can also work with this problem formulation. 
However, PPO outperforms other alternatives in terms of speed and quality of solutions found. In addition, PPO  is sample efficient which is very important in the history matching problem due to the fact that the objective function computation is costly.\\
In order to find multiple different solutions, it is important that we use a stochastic policy when we interact with the environment. Specifically, a stochastic policy will lead to a better exploration of the state-action space and hence it will find multiple solutions in each run.\\
The deep neural network design for the actor network ${\pi_{\theta}}$ is shown in Figure \ref{Fig:ActorNet}. It is composed of an input layer containing 27,000 neurons (equal to the number of uncertain parameters in SPE9); two hidden layers where each layer contains 4096 tanh activated neurons; and an output layer that is also equal to 27,000 as it needs to take action against 27,000 different uncertain values. The critic network $V(\theta)$, as shown in Figure \ref{Fig:CriticNet}, has the same structure as the actor network except for the last layer where it has only one neuron since it is only trying to estimate a scalar value.
\\
The second data set uses the same structure. However, since the problem has much fewer uncertain parameters (900 compared to 27,000), it uses in 900 neurons in the input and output layers. In addition, the number of the hidden neurons in the hidden layer is equal to 128 instead of 4096 used in the first data set.

\subsection{Parallel Reinforcement Learning History Matching}
\begin{figure}
    \centering
    \includegraphics[width=14cm]{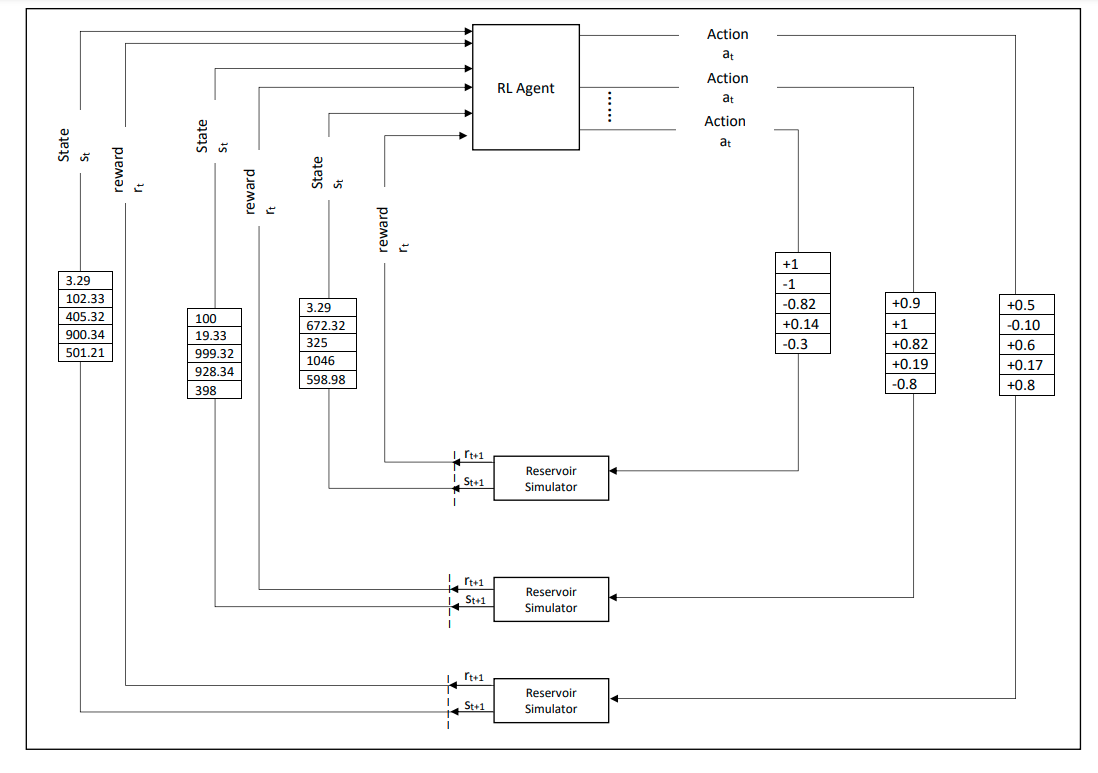}
    \caption{Redesigning the environment from Figure \ref{Fig:RSinRL} in order to enable the agent to learn from multiple simultaneous environments. The agent can observe $N$ states, take $N$ actions and collect $N$ rewards in parallel using the new architecture.}
    \label{Fig:ParallelRSinRL}
\end{figure}

History matching in general is a serial problem where the use of optimization methods to find the minimum of the objective function via gradient descent makes it necessary to know the previous objective function before taking the next step. However, in reinforcement learning the agent learns how to find the minimum by interacting with the environment one step at a time and collecting these experiences then training the deep neural network agent. So in essence, the deep neural network is trying to map states into actions so the agent needs only the current state, the action taken and the reward assigned to this action in order tune its deep neural network weights and choose the decisions that will maximize the rewards. 
PPO allows for the experiences to be collected into a batch every few time-steps and then sent to the agent for training even if the episode is not completed.\\
The way in which the actor-critic reinforcement learning train and optimize its agent gives us the chance to speed up the history matching process by training the agent in parallel and allowing it to collect experiences from running multiple scenarios of the history matching process at once. This can be done by allowing the agent to simultaneously interact with $N$ number of environments, observe different $N$ states, take $N$ number of different actions and collect $N$ different rewards as shown in Figure \ref{Fig:ParallelRSinRL}. For example, in the case of SPE9, the batch size is set to 192 where the algorithm will run 192 experiments by interacting with the environment then collects these experiments in a batch then sends it to the agent to train its deep neural networks and update their weights. Instead of waiting for a single environment to collect 192 experiments, the algorithm will launch eight environments in parallel where each environment collects 24 experiments which allows the agent to collect the 192 experiments in a much shorter time period.\\
Stable Baselines 3~\citep{stable-baselines3} implementation provides a wrapper that allow for the launch $N$ CPU threads, each thread works on an environment to collect the experiences then these experiences are sent to the GPU to train the agent's  deep neural networks in a faster manner. 
In addition to running multiple environments in parallel, inside each thread running an environment to collect the experiences, the environment itself can run the reservoir simulator in parallel to speed up the process of computing the objective function which is very common in reservoir simulation. For example, when running the experiment on SPE9, 4 MPI processes were used per environment.\\
However, it is very important to note that only the process of collecting the experiences and running the reservoir simulator can be done in parallel but each environment will run each episode in a serial manner and thus each environment must have its own desk directory with its own DATA, ECL and SMSPEC files. For example, the state of environment 12 at time-step ($t$) will depend on the state and the action of environment 12 at time ($t-1$) as this process {can not be parallelized}. The details of the algorithm is shown in Algorithm \ref{theParallelAlgorithm} pseudocode.

\begin{algorithm}
\caption{Parallel Reinforcement Learning History Matching} 
\label{theParallelAlgorithm}
\SetKwBlock{DoParallel}{do in parallel}{end}
	Launch N parallel environments \\
	Create active disk directory for each environment\\ 
	Set B = Batch Size\\
	Set t = 0\\
	Previous Objective Function $\mathcal{F}_{t-1}$ = Initial Objective Function $\mathcal{F}_{t=0}$\\
	\While {$t < $ max time-steps}{
		\DoParallel{
		\For {$i=0,1,\ldots$, $i < B/N$}{
		    Observe state $s_t$ \\
		    Take Action $a_t$ against uncertain parameters\\
		    Run Reservoir Simulator with adjusted parameters\\
		    Compute New Objective Function Value $\mathcal{F}_t$ \\
		    Reward  $r_t =  \mathcal{F}_{t-1} - \mathcal{F}_{t}$ \\
		    $\mathcal{F}_{t-1} = \mathcal{F}_t$ \\
		    \If{\space \space $\mathcal{F}_t$ $< \epsilon$ \space \space \space }{
               save history matched reservoir model to disk\\
            } 
		   }
		   }
	 Update  $\pi_{\theta}$ \& $V$  networks using $B$ experiments \\
    }
\end{algorithm}

\subsection{Reproducibility}
One of the major challenges faced while dealing with deep neural networks is reproducibility~\citep{Alahmari_2020,Huston_2018} in which repeating the experiment may lead to a different result with a different run-time taken to find the solution. The reproducibility problem is caused by the random network weights initialization prior to training in addition to using stochastic optimizers to optimize such weights. Another factor contributing to the reproducibility problem is the stochastic policy used by PPO  where a small random noise is added to the action suggested by the agent or in some cases some  the actions taken by the agents are sampled from a random noise matrix to ensure that the agent explores the environment efficiently to find multiple solutions to the history matching problem.\\
The reproducibility problem does not affect the outcome of the results in the first data set as all the solutions found meet the tolerance criteria. However, it becomes important when running the scalability test on the second data set due to the fact that some runs might find a solution quickly and other runs may take a while to find a solution. For this reason, an approach similar to~\citet{Alolayan_2021} was used to reduce the effects of the reproducibility problem where each of the scalability test run-time measurement is repeated 10 times and the average of these 10 runs is considered the final result.

\section{Results}\label{section:Results}
\begin{figure}
    \centering
    \includegraphics[width=14cm]{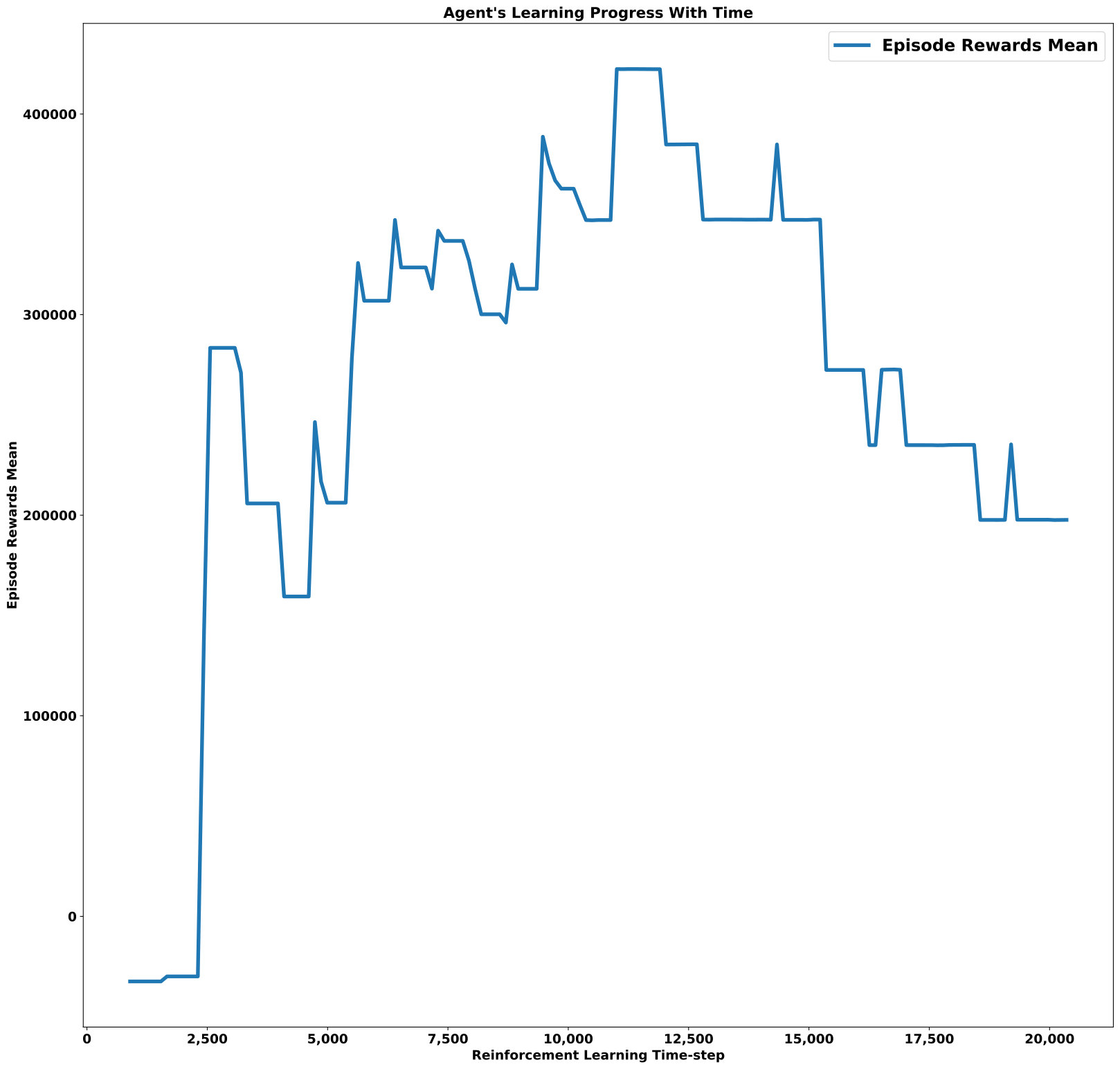}
    \caption{In the beginning of the training, the agent explore the environment often making bad actions and collecting negative rewards. Then, it starts to learn how to increase its rewards which enables it to find solutions that meet the tolerance criteria. }
    \label{Fig:RewardMean_50k}
\end{figure}

\begin{figure}
    \centering
    \includegraphics[width=14cm]{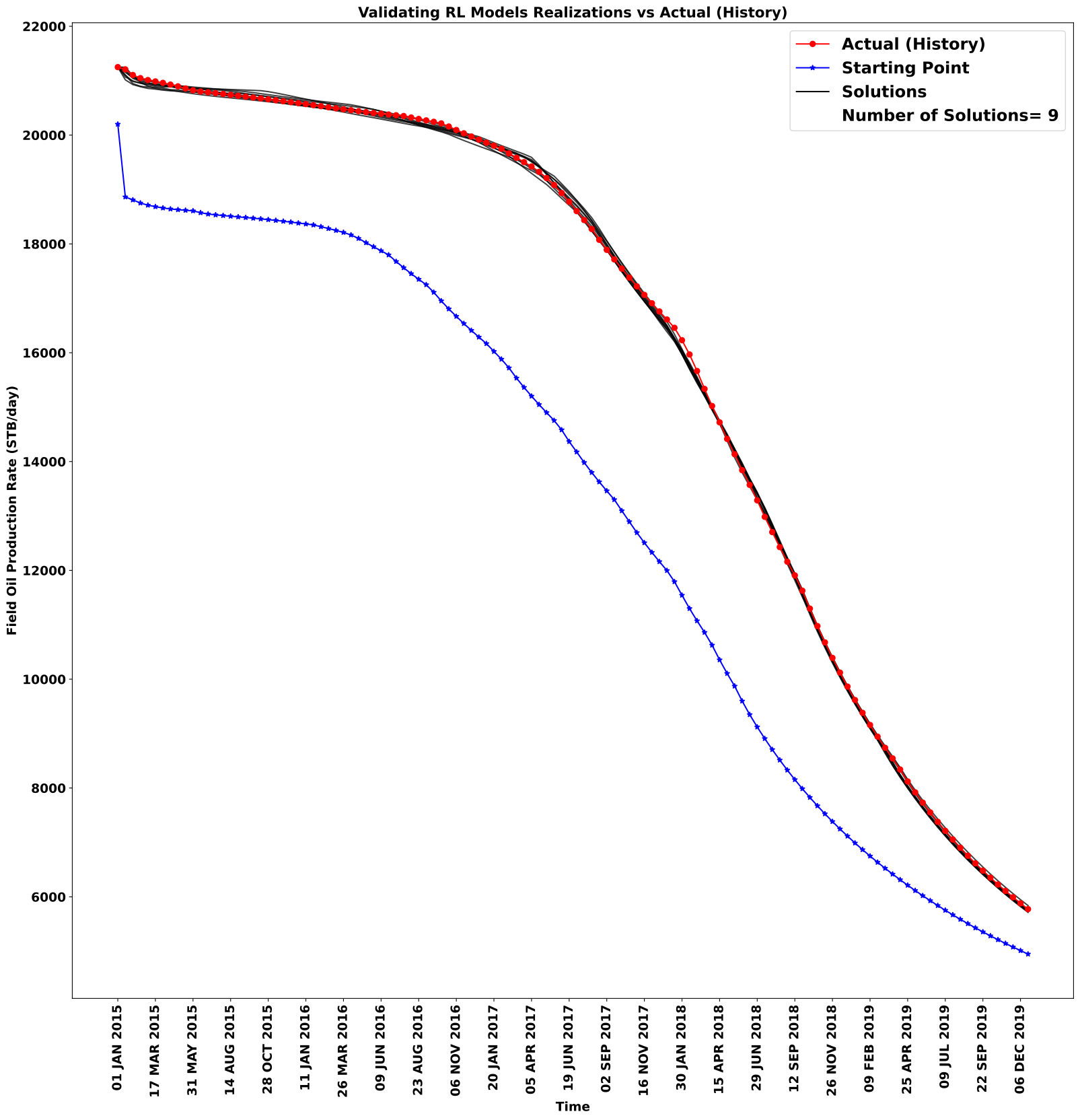}
    \caption{Algorithm \ref{theParallelAlgorithm} enabled the artificial deep neural network agent to learn from multiple environments simultaneously. { Thanks to the stochastic policy used, 9 multiple and different solutions to the history matching problem are found to the history matching problem during the 20,000 time-step.} }
    \label{Fig:multipleSols}
\end{figure}

\begin{figure}
    \centering
    \includegraphics[width=14cm]{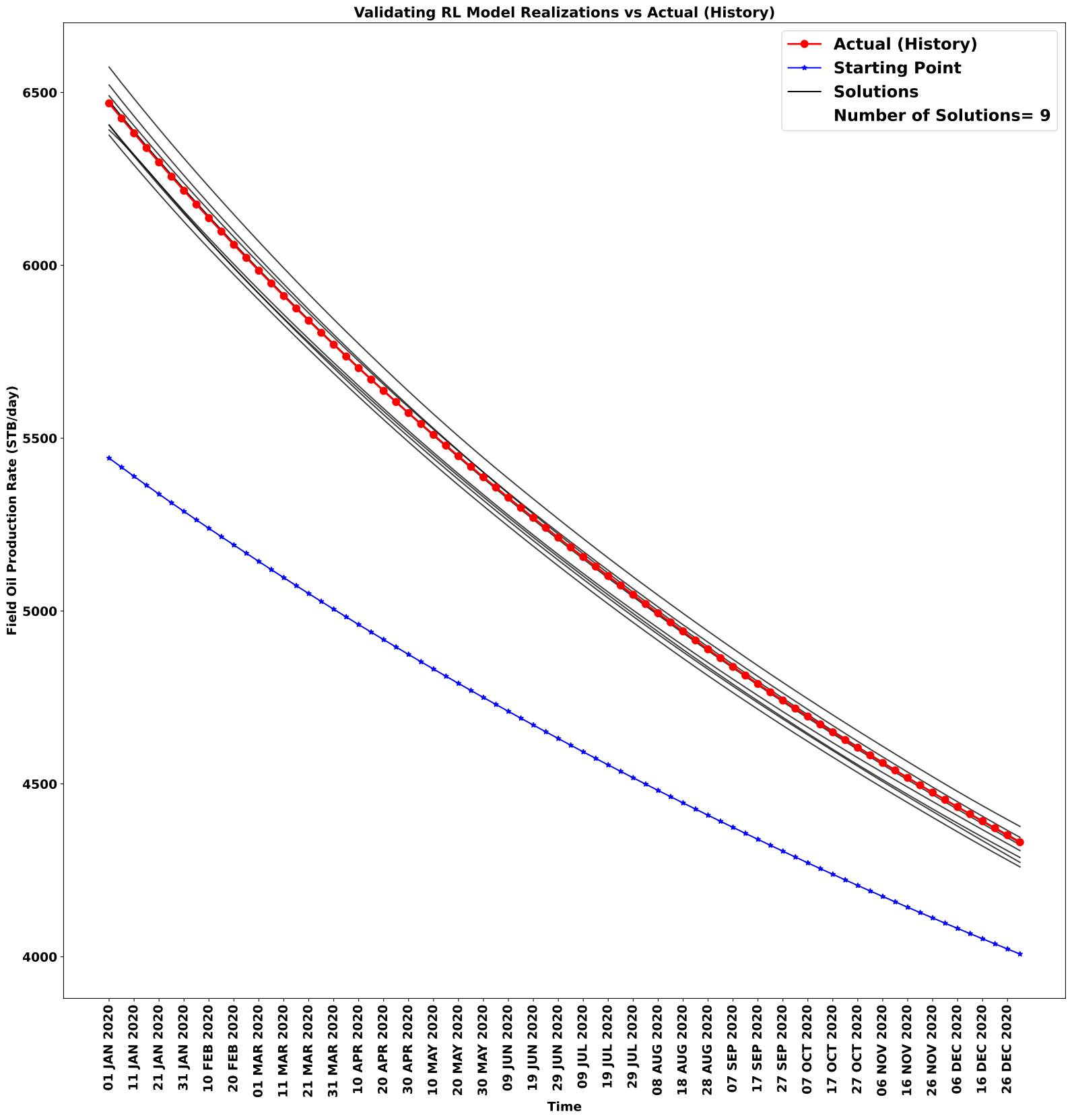}
    \caption{To test the forecasting capabilities of the realizations found by the artificial deep neural network agent, the realizations are validated by running them for one more year that the agent did not train on. The realizations found by the agent shows a good forecasting capabilities and provided multiple production scenarios that are within close range of the truth model. }
    \label{Fig:validation}
\end{figure}

\begin{figure}
    \centering
    \includegraphics[width=14cm]{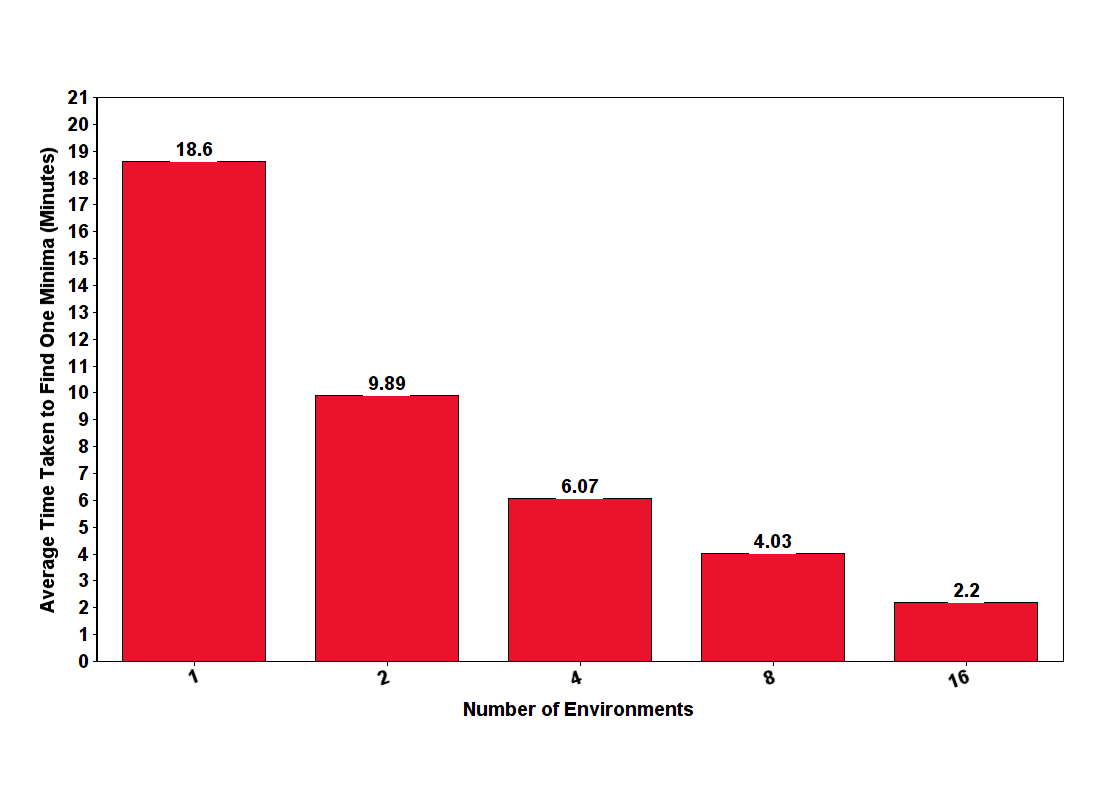}
    \caption{A significant reduction in run-time is achieved when computing resources are doubled. On average, a reduction of 41\%  in run-time is achieved when the computing resources are doubled resulting in total run-time reduction of 88\% when using 16 environments instead of one. For reproducibility purpose, each column represents the average time taken across 10 runs to find one solution to the history matching problem.}
    \label{Fig:Scalability}
\end{figure}

\begin{figure}
    \centering
    \includegraphics[width=14cm]{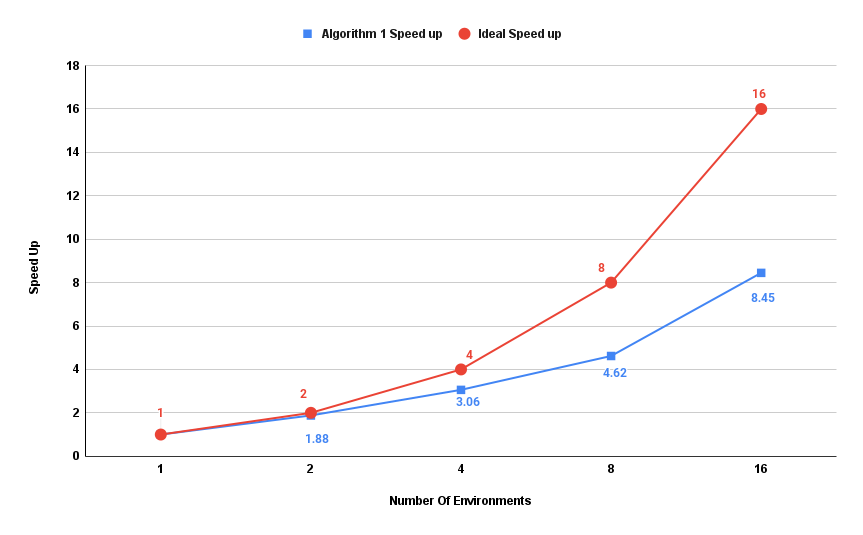}
    \caption{Based on the run-time values shown in Figure \ref{Fig:Scalability}, Algorithm \ref{theParallelAlgorithm} scalability plot shows a significant speed up when more computing resources are added. On average, a speed of 1.57 is achieved every time the computing resources are doubled.}
    \label{Fig:speed_up}
\end{figure}

\begin{figure}
    \centering
    \includegraphics[width=14cm]{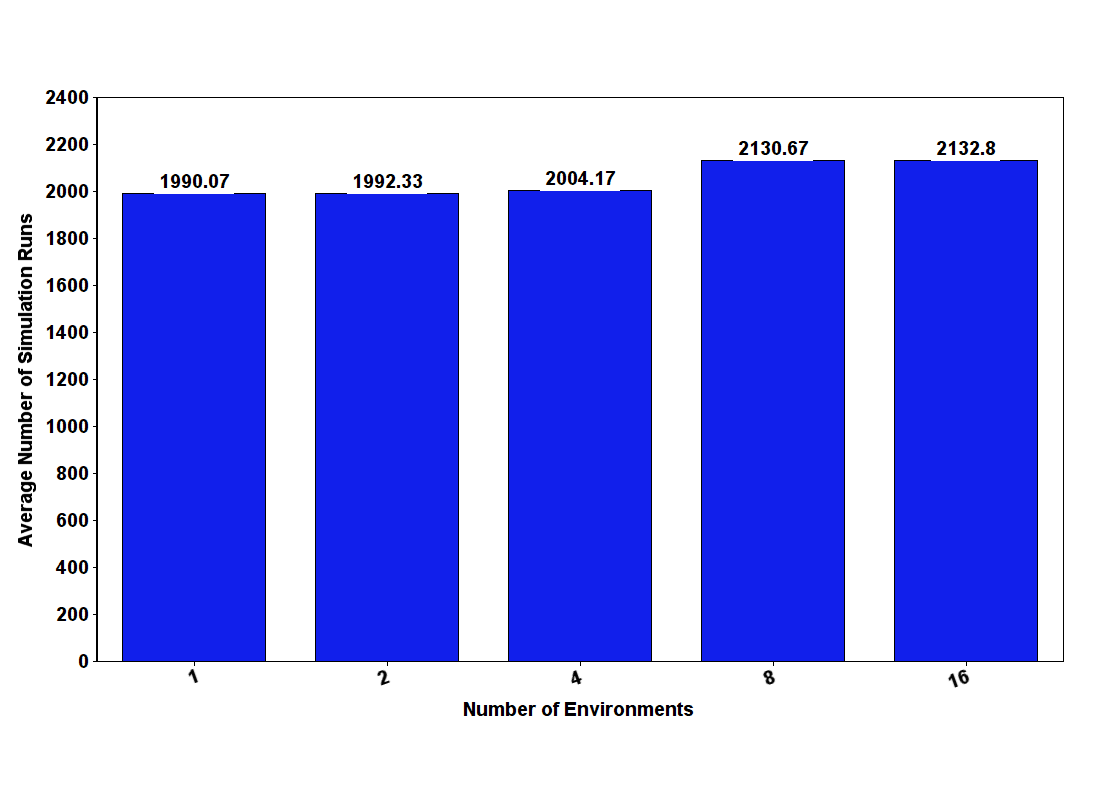}
    \caption{Based on the run-time values reported in Figure \ref{Fig:Scalability}, Algorithm \ref{theParallelAlgorithm} had the capacity to efficiently scale when increasing the number of environments without significantly increasing the number of simulation runs. }
    \label{Fig:NumberOfRuns}
\end{figure}

Algorithm \ref{theParallelAlgorithm} was used {to run 20,000 reinforcement learning time-steps in parallel} on the first data set to search the parameters space for solutions to the history matching problem. As shown in Figure \ref{Fig:RewardMean_50k}, in the beginning of training the algorithm behaved as expected from any reinforcement learning algorithm where it kept making wrong decisions and collecting negative rewards. As the agent spends more time interacting with the environment, it learns how to map states into actions that enables it to increase its rewards.
\\
Per the definition of the reward function in Equation \ref{RewardFunc}, as the agent tries to increase its rewards, it will tune the uncertain parameters and find models that reduce the objective function. As the agent spends more time in the environment trying to maximize its reward, it will reduce the objective function to meet the tolerance criteria and thus it will find a solution to the history matching problem. As shown in Figure \ref{Fig:multipleSols}, thanks to the stochastic policy the agent uses, it can find multiple different solutions as it tries to take different trajectories to better explore the environment. \\
The approach of using a stochastic policy may encourage the agent to take bad actions and waste computing power. However, it is necessary for the agent to do so in order to search for new different solutions as this is always the case with the exploration-exploitation dilemma.  Such behaviour can be seen from the agent's learning progress it appears in Figure \ref{Fig:RewardMean_50k} around the 13,000 time-step. Although the agent found few solutions and collected positive rewards by that time-step, it did not repeat its actions to maximize rewards. Instead, the agent explored the environment for new different solutions causing it to make decisions that resulted in less rewards than previously acquired.
\\
{In order to study the forecasting capabilities for the model realizations found by the artificial deep neural network agent, the quality of the models are then validated by running each model for an additional year that the agent did not train on. Figure \mbox{\ref{Fig:validation}} shows each realization forecast compared to the truth model and the starting point. As shown from the figure, the realizations found by the agent exhibited a good forecasting capabilities for the reservoir model and provided multiple production scenarios where some models over-predicted and some models under predicted. This can help engineers run uncertainty quantification analysis with a higher degree of confidence. When using this algorithm, it is recommended to reserve a portion of the historical data for validation in order to assess the quality of the solutions found by the agent. Additionally, the validation period can also be used by the engineers to further filter out unwanted solutions based on their own criteria.}
\\
To examine and measure how Algorithm \ref{theParallelAlgorithm} scales with the number of available computing resources, a scalability test was conducted on the second data set. The scalability test was conducted using the second data set as it has a fewer number of uncertain parameters and a smaller initial objective function, making the problem easier to solve resulting in a shorter run-time. A reasonable run-time is necessary for the scalability test as it requires the experiments to be repeated tens of times for reproducibility purpose as discussed in the Methodology. 
Figure \ref{Fig:Scalability} shows the capacity of Algorithm \ref{theParallelAlgorithm} to reduce the run-time when doubling the number of available computing resources. On average, every time the computing resources are doubled, a 41\% reduction in run-time is achieved. When using 16 environments instead of one, Algorithm \ref{theParallelAlgorithm} reduced the total run-time from 18.6 minutes to 2.2 minutes, achieving an 88\% reduction in the time needed to find one solution. 
\\
As shown in Figure \ref{Fig:speed_up}, Algorithm \ref{theParallelAlgorithm} achieved a speed up of 8.45 when using 16 environments compared to one environment. The algorithm scored an average speed up of 1.57 when doubling the number of resources compared to a speed up of 2 in the ideal scenario where the ideal scenario refers to the non-realistic case when an algorithm achieves a speed up of 2 every time the resources are doubled.  The algorithm achieved a maximum speed up of 1.88  when running the algorithm on two environments instead of one, and a minimum speed up of 1.51 when running the algorithm on eight environments instead of four. 
\\
{ A major advantage of the parallelization procedure employed by Algorithm \mbox{\ref{theParallelAlgorithm}} is the fact that the number of reinforcement learning time-steps (number of forward simulation runs) did not increase significantly while increasing the number of computing resources as shown in Figure \mbox{\ref{Fig:NumberOfRuns}}. Such property indicates that the algorithm can scale efficiently when adding computing resources. As shown in the figure, it takes the algorithm around 2,000 time-steps to find a minima and instead of taking 18.6 minutes running on one environment, it only took 2.2 minutes when running on 16 environments while slightly increasing the number of simulation runs from around 1990 to around 2133 runs.}

\section{Discussion}\label{section:Discussion}

\begin{figure}
    \centering
    \includegraphics[width=15cm]{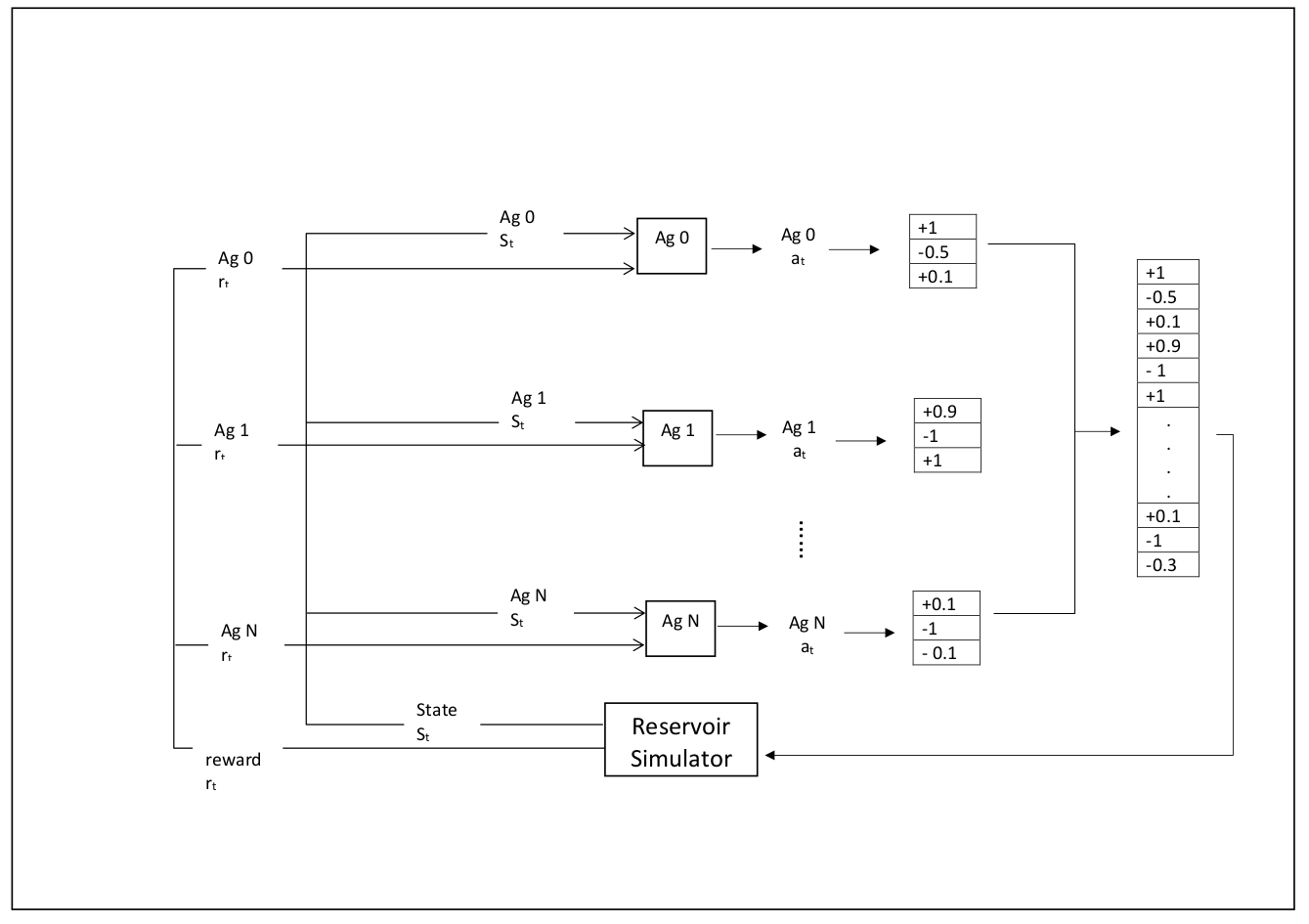}
    \caption{A multi-agent approach to solve the history matching problem. Using multiple collaborating artificially intelligent agents may provide a mechanism to handle more complex problems.}
    \label{MultiAgent}
\end{figure}

By reformulating the history matching problem from a mathematical least-square optimization problem into a Markov Decision Process problem, the suggested algorithm provided a way in which the history matching problem can be solved in parallel where adding more computing resources can speed up the convergence process. In addition, the suggested algorithm can be used to find multiple and different solutions to the history matching problem, allowing for better forecasting uncertainty analysis. By creating multiple environments and allowing the artificial neural network agent to sample from these multiple environments simultaneously, the algorithm had the capacity to tune 27,000 uncertain parameters. Such capacity allowed the agent to find multiple and different historically matching models in a timely manner.\\
The multiple different solutions shown in Figure \ref{Fig:multipleSols} does not have to necessarily be 9 solutions as more solutions can be found. As a matter of fact, there are infinite solutions to the problem \citep{LiAndRuijian_2003} and if more solutions are desired to better assess the uncertainty, then the engineer running the experiment can run Algorithm \ref{theParallelAlgorithm} for a longer period of time. Giving the agent more time to learn from the environment will allow it to find more solutions. The process of training the agent for a longer time to find more solutions does not require the training process to be repeated from scratch. {In order to avoid retraining, the agent's knowledge can be re-used by saving the deep neural network models to the disk and then re-loading them later to resume training if the number of solutions found is not sufficient.}\\
The results also show that the agent learns how to interact with the environment in order to maximize its rewards, where in the beginning the algorithm will not be able to find any solutions until it learns how to map the current states of the environment into actions that reduce the objective function,  eventually finding a solution. Naturally,  the more complex the problem (i.e., the more uncertain parameter needs to be tuned) the longer the training needed. This is due to the fact that the agent will need more samples to explore and learn from the environment. In addition, the deep neural networks used to map the state into action will be larger as explained in the Methodology, thus needing more data to be able to successfully find multiple solutions to the history matching problem.\\
{It is important to also note that as the number uncertain parameters increases, the training process becomes more difficult. Such difficulty is expected as the higher the number of uncertain parameters, the larger more complex the deep neural network needed to handle the problem. A larger network can cause instability in training and would require a smaller learning rate that slows down the algorithm. We found out that adding a regularization term to the objective function would not only help mitigate the ill-posedness effect on the problem, but it also can help stabilize the learning process and allow for a larger learning rate to be used which can speed up the convergence of the algorithm. }\\
Not only can the number of parameters  affect the speed of convergence, the starting point also plays an important rule. Naturally, a model with a small initial objective function of 100,000 will converge faster than a model with an initial objective function of 500,000. This is an expected behavior that also occurs when using optimization algorithms to solve the problem where the closer the starting point to the solution, the faster the convergence. {Faster convergence also can be achieved if the tolerance criteria is loosened as this is the case with other algorithms as well \mbox{\citep{Heidari2011}}. Reducing the historical mismatch by more than 99\% when tuning 27,000 uncertain parameters will always be difficult as it will require most algorithms to get more samples from the reservoir model parameters. As with other algorithms, a trade-off between accuracy and speed will also needs to be considered. Moreover, faster convergence may also be achieved if the maximum time-steps per episode parameter is set to a smaller value. However, this might hinder the algorithm's ability to explore the parameters space efficiently as sometimes the agent takes few bad actions on purpose in order to search for new solutions.}\\
{ The run-time might be reduced by reducing the number of uncertain parameters using sensitivity analysis where the cells that have the least effect on the change of the objective function can be ignored and not included in the uncertain parameters vector just like when dealing with inactive cells. The number of uncertain parameters can also be reduced using permeability multipliers, where instead of tuning each single permeability value, a group of constant multipliers are tuned, where each multiplier is used to multiply the permeability values in certain cells. However, in both data sets used in this research paper, no multipliers are used in order to show the capacity of the algorithm to tune thousands of parameters.
Another approach to reduce the run-time is the use of proxy models\mbox{~\citep{He_2016,Shams_2017,Negash_2016}} where the reservoir simulator $f{(u)}$ is replaced by a fast approximate model $g{(u)}$ that can map the input of the reservoir simulator into an output within some acceptable accuracy.}\\
{A major advantage of using the suggested algorithm is its flexibilty in terms of the parameters space. Where unlike Ensemble Kalman Filter (EnKF), it {does not} require the parameters space to have a Gaussian distribution.
Another advantage is its versatility, where some optimization algorithms might diverge away from the solution if the initial objective function is large, this algorithm is less prone to divergence with an initial guess that is reasonably far from any solution if given enough time. This is mainly caused by the nature of this algorithm where it would restart itself to the starting point upon divergence. Then, thanks to the shaping of the reward function the agent will be punished severely for diverging. By punishing the agent for diverging away from the solution, it will reduce the probability of choosing these actions in the future to enable the agent to learn from such mistakes and not repeat the same actions again that lead to divergence.}\\
It is important to note that when using a very large number of computing resources across hundreds of environments, { the algorithm scalability will be lower compared to the case when utilizing tens of environments }. This is due to the fact that the batch size (1024, 512, etc.) may not be big enough to divide the work efficiently amongst the environments. When this occurs, the batch size would have to be fixed when increasing the number of environments causing PPO to be less sample efficient as resources increase. In that case, a good speed up is still expected to be achieved when adding more computing resources but Algorithm \ref{theParallelAlgorithm} { might not scale very well compared to its scaling performance when using tens of environments.} When dealing with a very large model that requires hundreds of environments, engineers will have to spend some time adjusting the hyperparameters and ensuring efficient workload for each environment in order to achieve high scalability.\\
One of the major challenges arising from using Algorithm \ref{theParallelAlgorithm} is the number of hyperparameters related to each experiment. Where in addition to the usual deep neural networks hyperparameters (batch size, learning rate, number of neurons, etc.) and reinforcement learning hyperparameters ($\gamma$, maximum time-steps, etc.), the experiments itself has its own hyperparameters, such as $K_\Delta$ and maximum time-steps per episode as these parameters are problem-specific that depend on the reservoir model at hand. The engineer using the suggested algorithm might have to spend some time trying to find some optimal parameters that can make the algorithm converge faster. \\
The vector containing all uncertain parameters ${u}$ does not have to only contain permeability values. Any uncertain parameters, for example the porosity $\phi$, can also be included in ${u}$. However, the engineer running the experiment must also adjust the action space $\mathcal{A}$ to take into account the scale of change between different properties at each reinforcement learning time-step, where the maximum change allowed for the permeability value $K_\Delta$ might differ than the maximum change for porosity $\phi_\Delta$. \\
A future research opportunity can be explored by extending this framework using Multi-Agent Reinforcement Learning (MARL)~\citep{Zhang_2021,Hoen_2005,Panait_2005} instead of using a single agent across multiple environments. As shown in Figure \ref{MultiAgent}, the multiple agents can work collaboratively towards reducing the objective function where each agent is responsible for a section of the reservoir model. Each agent can observe the entire environment or just a part of it to take actions. Such approach can allow researchers to tackle very large models because each agent would have to map an easier function with a smaller number of actions. The small number of actions per agent would result in smaller deep neural networks, requiring less training time.
\\
\noindent However, such an approach comes with its own difficulties as well, where it would be difficult to come up with an appropriate reward function. The most difficult part would be figuring out a way to reward good actions and punish bad ones, where at each time-step it would be difficult to identify which agents took good actions and which agents took bad ones from a single reward function. 

\section{Conclusions}\label{section:Conclusion}

The results drawn from this research paper show that suggested parallel automatic history matching algorithm  using reinforcement learning (Algorithm \ref{theParallelAlgorithm}) can train an artificial deep neural network agent that is capable of finding multiple different solutions to the history matching problem, even when dealing with tens of thousands of uncertain parameters. By reformulating the problem from an optimization problem into a Markov Decision Process, the algorithm gave the chance to sample data from multiple trajectories across multiple environments, allowing the agent to learn faster as more computing resources are added.\\
As shown from the Results, the algorithm achieved an average speed up of 1.57 when the computing resources are doubled and had the capacity to reduce the run-time needed to find one solution by 88\% from 18.6 minutes to 2.2 minutes, when using 16 environments instead of one. Such parallelization gave the opportunity to tackle complex problems and find multiple solutions in a timely manner when tuning 27,000 uncertain parameters.

\section*{Acknowledgments}
{The authors would like to thank Dr. Ali Dogru and Dr. Tariq Alshaaln from Saudi Aramco EXPEC Advanced Research Center as well as Dr. Ruben Juanes, Dr. Herbert Einstein and Mohammed Alsobay from Massachusetts Institute of Technology
for their helpful advice and support for this research. The authors would also like to thank \href{https://opm-project.org/}{Open Porous Media Initiative} for providing open source reservoir simulator and data sets that were used to test this algorithm. Omar S.  Alolayan would like to thank Saudi Aramco for their graduate fellowship support.}

\section*{conflicts of interest}
{The authors declare that they have no known competing financial
interests or personal relationships that could have appeared to influence the work reported in this paper.} 
\section*{{Nomenclature}} \label{section:Nomenclature}
$a_t$     : Reinforcement learning action taken by an agent at time-step $t$.\\
$\mathcal{A}$      : Set of all possible actions that can be taken by the agent.\\
$Ag$ : artificial deep neural network agent.\\
$B$: Reinforcement learning batch size.\\
$f{(u)}$: Function representing the reservoir simulator. \\
$g{(u)}$: Function representing the proxy model.\\
$K_x$,$K_y$ and  $K_z$ : Permeability in the x,y and z directions.\\
$K_\Delta$: Permeability limits of each action taken.\\
$m$ : Number of time-steps in the simulation model.\\
$n$ : Number of wells in the reservoir model.\\
$N$: Number of environments.\\
$q$ : Actual pressure or saturation from historical data.\\
$\hat{q}$ : Simulated rate from the simulator or proxy model.\\
$r_t$     : The reward obtained from the environment at time-step ${t}$.\\
$s_t$ : Current state of the reinforcement learning environment at time-step ${t}$.\\
$\mathcal{S}$ : Set of all possible states in the reinforcement learning environment.\\
$t$      : Reinforcement learning time-step iterator.\\
${u}$ : Vector containing all uncertain parameters in the model.\\
$X$,$Y$ and $Z$: Number of cells in the x,y and z direction in the reservoir model.\\
$\epsilon$ : Error tolerance level accepted.\\
$\gamma$: Reinforcement learning discount factor.\\
${\pi_{\theta}}$: PPO Actor Network.\\

\section*{\textbf{Appendix}}
\textbf{Experiments Hyperparameters}\\
\noindent $K_\Delta$ = 100 for for PermX, PermY and PermZ in SPE1.\\
$K_\Delta$ = 100 for PermX and PermY and = 1 for PermZ in SPE9.\\
$\alpha$ = $1e^{-3}$ \\
$C_{q}$ = Identity Matrix (I).\\
$C_{u}$ = Identity Matrix (I).\\
$\lambda$ = 1 for SPE9 and = 0.1 for SPE1. \\
Algorithm total time-steps = 20,000 for SPE9 and until at least 1 solution is found for SPE1.\\
Learning rate = $1 \times 10^{-5}$ for SPE9 and = $9 \times 10^{-4}$ for SPE1.\\
MPI Processes per environment = 4 for SPE9 and = 1 for SPE1. \\
OMP threads per MPI Process = 2.\\
$\gamma$ = 0.99. \\
Tolerance $\epsilon$ = 1,000 for SPE1 and = 2,500 for SPE9.\\
Reward for good termination = $1 \times 10^{4}$ for SPE1 and = $2.5 \times 10^6$ for SPE9.\\
Reward for divergence = $-1\times 10^4$ for SPE1 and = $-2.5 \times 10^6$ for SPE9.\\
Reward for exceeding maximum time-steps per episode = $-1\times10^4$ for SPE1 and = $-1.25\times10^6$ for SPE9.\\
PPO Clipping Range = 0.2.\\
Maximum time-steps per episode = 100.\\
Batch size = 192 for SPE9 and = 32 for SPE1. \\

\newpage
\clearpage
\bibliography{refs}

\end{document}